\documentclass{article}

\usepackage{microtype}
\usepackage{graphicx}
\usepackage{booktabs} %

\usepackage{hyperref}

\usepackage[linesnumbered,ruled,vlined,norelsize]{algorithm2e}

\usepackage[accepted]{icml2024}

\usepackage{amsmath}
\usepackage{amssymb}
\usepackage{mathtools}
\usepackage{amsthm}
\usepackage{booktabs}
\usepackage{multirow}
\usepackage{booktabs}
\usepackage{multirow}
\usepackage{colortbl}
\usepackage{dsfont}
\usepackage[capitalize,noabbrev]{cleveref}

\theoremstyle{plain}

\theoremstyle{definition}

\theoremstyle{remark}

\usepackage[textsize=tiny]{todonotes}

\icmltitlerunning{Localizing Task Information for Improved Model Merging and Compression}
\usepackage[short]{optidef}

\usepackage{my_commands}

\usepackage{paralist}
\usepackage{thmtools}
\usepackage[most]{tcolorbox}

\newcommand{\draftfootnote}[1]{\footnote{}}
\tcbset{%
    theo/.style={%
        enhanced,
        breakable,
        sharp corners,
        boxed title style={empty, size=minimal, bottom=1.5mm, },
        toprule=0pt, rightrule=0pt, bottomrule=0pt, leftrule=1mm,
        colback=#1!5, colframe=#1!80!black, coltitle=#1!80!black, 
        detach title,
        overlay unbroken and first ={
        
        }
    }
}

\newtcbtheorem[auto counter]{thquestion}{Question}{theo=blue}{th}

\begin{document}

\newcommand{\foo}[2]{$\text{#1}\ _{(\text{#2})}$}
\newcommand{\myparagraph}[1]{\textbf{#1} }
\twocolumn[
\icmltitle{
Localizing Task Information for Improved Model Merging and Compression
}

\icmlsetsymbol{equal}{*}

\begin{icmlauthorlist}
\icmlauthor{Ke Wang}{equal,epfl}
\icmlauthor{Nikolaos Dimitriadis}{equal,epfl}
\icmlauthor{Guillermo Ortiz-Jiménez}{deepmind,whileepfl}
\icmlauthor{Fran\c{c}ois Fleuret}   {unige}
\icmlauthor{Pascal Frossard}{epfl}
\end{icmlauthorlist}

\icmlaffiliation{epfl}{École Polytechnique Fédérale de Lausanne, Lausanne, Switzerland}
\icmlaffiliation{deepmind}{Google Deepmind}
\icmlaffiliation{whileepfl}{Work done while at EPFL.}
\icmlaffiliation{unige}{University of Geneva, Geneva, Switzerland.}

\icmlcorrespondingauthor{Ke Wang}{k.wang@epfl.ch}
\icmlcorrespondingauthor{Nikolaos Dimitriadis}{nikolaos.dimitriadis@epfl.ch}

\icmlkeywords{Machine Learning, ICML}

\vskip 0.3in
]

\printAffiliationsAndNotice{\icmlEqualContribution} %
\newcommand{\pretrained}{pre-trained\xspace}
\newcommand{\conditionalclearpage}{\clearpage}
\renewcommand{\conditionalclearpage}{}
\newcommand{\algoname}{\texttt{XYZMASKS}\xspace}
\newcommand{\secondalgoname}{\texttt{Consensus Task Arithmetic}\xspace}
\renewcommand{\secondalgoname}{\texttt{Consensus Merging}\xspace}
\renewcommand{\secondalgoname}{Consensus Merging\xspace}
\newcommand{\firstalgoname}{\texttt{Specialized Task Arithmetic}\xspace}

\renewcommand{\algorithmautorefname}{Algorithm}
\newcommand{\interference}{\textit{task interference}\xspace}
\newcommand{\maskname}{\textit{mask-algo-name}\xspace}
\renewcommand{\maskname}{\texttt{TALL-masks}\xspace}
\newcommand{\mtv}{\textit{multi-task vector}\xspace}
\newcommand{\nuisance}{\textit{nuisance weights}\xspace}
\newcommand{\selfish}{\textit{selfish}\xspace}
\newcommand{\catastrophic}{\textit{catastrophic}\xspace}
\newcommand{\ties}{TIES\xspace}
\renewcommand{\subsectionautorefname}{Section}

\begin{abstract}  
Model merging and task arithmetic have emerged as promising scalable approaches to merge multiple single-task checkpoints to one multi-task model, but their applicability is reduced by significant performance loss. Previous works have linked these drops to interference in the weight space and erasure of important task-specific features. Instead, in this work we show that the information required to solve each task is still preserved after merging as different tasks mostly use non-overlapping sets of weights. We propose TALL-masks, a method to identify these task supports given a collection of task vectors and show that one can retrieve $>99\%$ of the single task accuracy by applying our masks to the multi-task vector, effectively compressing the individual checkpoints. We study the statistics of intersections among constructed masks and reveal the existence of \textit{selfish} and \textit{catastrophic} weights, i.e., parameters that are important exclusively to one task and irrelevant to all tasks but detrimental to multi-task fusion. For this reason, we propose Consensus Merging, an algorithm that eliminates such weights and improves the general performance of existing model merging approaches. Our experiments  in vision and NLP benchmarks with up to 20 tasks, show that Consensus Merging consistently improves existing approaches. Furthermore, our proposed compression scheme reduces storage from 57Gb to 8.2Gb while retaining 99.7\% of original performance.
\looseness=-1

    
\end{abstract}

\section{Introduction}
\label{sec:intro}
{

In recent years, the field of ML has witnessed a paradigm shift with the release of foundation models and the influx of associated checkpoints, significantly improving the performance on downstream applications \citep{devlin-etal-2019-bert,ilharco2022patching,Wortsman_Ilharco_Kim_etal_2022,pruksachatkun-etal-2020-intermediate,zhou2023comprehensive}. The widespread adoption of foundation models has followed a proliferation of works addressing practical challenges arising from their sheer computational and storage requirements. For example, parameter-efficient fine-tuning \citep{hu2022lora,liu2022few}, quantization \citep{dettmers2022gptint,dettmers2023qlora}   address aspects of training, fine-tuning and inference. 
An important question remains how to efficiently leverage the existing fine-tuned models towards improving models and building generalist agents \citep{reed2022a}.

\begin{figure*}[t]
    \centering
    \includegraphics[width=\linewidth]{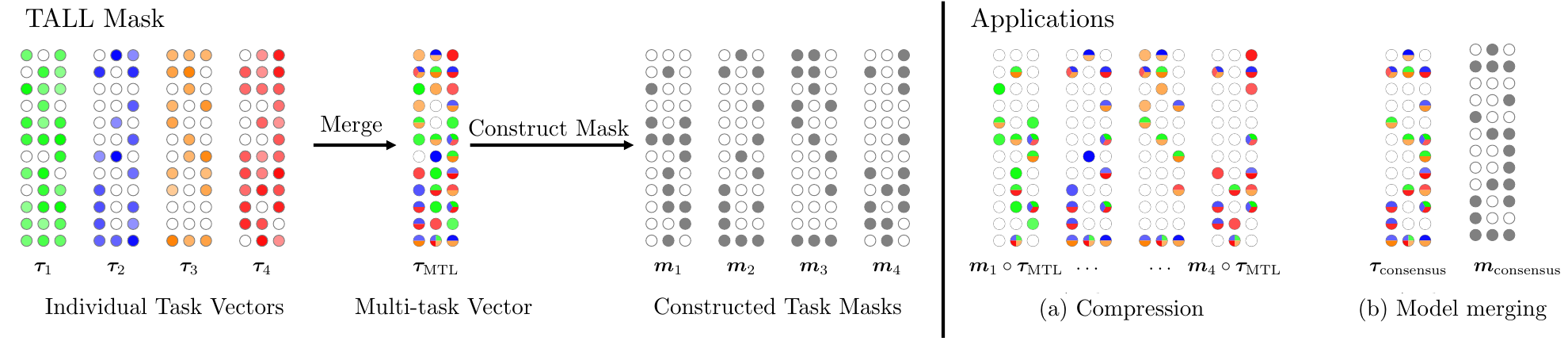}
        \caption{
    Illustration of our mask construction algorithm (left) along with the applications (right) on model compression and model merging. Each block corresponds to the same weight matrix, and color intensity reflects the value of each parameter -- empty means zero value. 
    Given single-task vectors $\{\tv{t}\}_{t=1}^4$ and the merged vector $\tvopt$, our method constructs per-task masks $\{\mm_t\}_{t=1}^4$, pinpointing the important  parameters for each original task vector. For \textit{model merging}, we keep only the  `\textit{general}' weights selected by more than one mask and produce the consensus mask $\mm_{\textrm{consensus}}$ and the final merged vector. For compression, we evaluate on each task with reconstructed task vectors by masking out the irrelevant weights, retaining almost full performance without saving the individual task vectors.
    }
    \label{fig:illustration}
\end{figure*}
Recent work has illuminated the benefits of interpolating the weights of different models \citep{frankle2020linear,ilharco2022patching,wortsman2022model,ortiz2023task,dimitriadis2023pareto}, offering scalable and effective techniques to edit  the knowledge of pre-trained models. 
Task arithmetic (TA) \citep{ilharco2023task} has emerged as a promising solution to fuse the knowledge of disparate checkpoints into a single model with  multi-objective capabilities, forgoing the need for additional joint training \citep{Caruana_1997} or optimizing over the exponentially large number of task combinations  \citep{Standley_Zamir_Chen_etal_2020,Fifty_Amid_Zhao_etal_2021}. Prior studies have proposed more involved merging techniques by resolving \textit{weight} interference \citep{Yadav_Tam_Choshen_etal_2023}, matching activations \citep{jin2023dataless} or by preserving task-specific important parameters \citep{matena2022merging,Tam_Bansal_Raffel_2023}.
Despite these recent advances, weight space interpolation for multi-task fusion still suffers from significant drops in performance compared to individual fine-tuned models.
\looseness=-1

In this paper, we present a novel view and show that performance of the merged model can degrade even without weight interference or information erasure through a controlled experiment. In contrast, the discriminant information for individual tasks is preserved and embedded in the multi-task vector after merging disparate task vectors, and we propose an algorithm, \maskname, that identifies the subset of important parameters for each task \citep{grafting,dai-etal-2022-knowledge,bayazit2023discovering}. We cast the problem of localizing important information as approximating each original task vector via erasing task-irrelevant information in the merged multi-task vector with a data-driven way, resulting in the construction of task-specific  binary masks. We study the statistics of mask agreements among tasks, and reveal the existence of \textit{catastrophic} and \textit{selfish} weights, i.e., parameters that are deemed important by none and exclusively one task, respectively.
\looseness=-1

We then propose \secondalgoname, a method 
that utilizes the constructed masks to eliminate the \textit{catastrophic} and \textit{selfish}  weights and is complementary to existing model merging approaches. Through extensive experimental validation, we show that our proposed \secondalgoname consistently improves prior methods. For instance, building upon task arithmetic \citep{ilharco2023task} yields 4.9\% gain in absolute average accuracy on a 20 task vision benchmark, while we improve \ties \citep{Yadav_Tam_Choshen_etal_2023} by 6.3\% on an 8-task NLP benchmark. 
\looseness=-1

We also employ the constructed masks towards compressing the individually fine-tuned checkpoints. Motivated by our findings that task-specific information is preserved and by virtue of the masks, we can localize the knowledge of each task in the merged vector and extract it to approximate the original single-task vector. We compress the collection of checkpoints to the zero-shot model, the merged task vector and binary masks. Our experimental validation shows that our algorithms retains $>99\%$ of original performance in various vision settings, ranging from small to large ViTs \citep{dosovitskiy2021an} and benchmarks from 8 to 20 tasks, showing remarkable robustness to the increase in number of tasks. For instance, in a vision benchmark we compress 20 fine-tuned models from 57Gb to 8.2Gb retaining $99.7\%$ of performance, while model merging methods almost reset to mere zero-shot performance.
\footnote{{The source code can be found at \url{https://github.com/nik-dim/tall_masks}}.}
\looseness=-1

In short, our contributions are the following:
\begin{itemize}
    \item We show that the task-specific information is preserved after merging, but task arithmetic cannot properly utilize it due to \textit{task interference}. We provide an efficient algorithm, \maskname, to localize the task-specific information in the multi-task vector, which deactivates irrelevant parts for each task in the merged multi-task vector with binary masks.\looseness=-1

    \item With the constructed task-specific masks, we are able to eliminate task interference and compress multiple fine-tuned checkpoints to only the zero-shot model, the merged task vector and the aforementioned binary masks while preserving the performance of individual models. 
    \looseness=-1
    
    \item We analyze the profile of mask agreements and identify the existence of weights deemed important by only one task or even none. We then propose \secondalgoname, a model merging method that eliminates these \selfish and \catastrophic weights, keeping only \textit{general} weights. Our method can be combined with existing approaches, such as Task Arithmetic or TIES, and consistently improve over them, showing better robustness to increasing number of tasks. 
    \looseness=-1
    
    \item We perform extensive evaluation on Computer Vision and NLP benchmarks and show the benefits of our proposed methods. For model merging, our \secondalgoname consistently improves prior merging methods, setting state-of-the-art results.
    For compression, we achieve $>99\%$ performance retention in across all vision benchmarks and model sizes while requiring much less storage compared to the original collection of fine-tuned checkpoints. 
    \looseness=-1
\end{itemize}

\section{Related Work}
\label{sec:related work}

\myparagraph{Weight Interpolation and Model Merging}
Model editing  directly in the weight space has attracted a lot of attention in recent years with many works showing that interpolating the weights of different models results in low-loss paths \citep{Garipov_Izmailov_Podoprikhin_etal_2018,Draxler_Veschgini_Salmhofer_etal_2018,frankle2020linear}. \citet{wortsman2021learning} enacted on these insights and trained a weight ensemble from scratch, showing better generalization \citep{Foret_Kleiner_Mobahi_etal_2021,Chaudhari_2018} for the midpoint, while \citet{dimitriadis2023pareto} extended these ideas to Multi-Task Learning and showed that linear weight subspaces can encode tradeoffs and map to the Pareto Front.
While these works focus on end-to-end training, \citet{ilharco2023task} studied \pretrained models and observed that arithmetic operations among fine-tuned weights generate similar functional responses and allow for a scalable framework to endow multi-objective capabilities.     
Several approaches have improved this idea by performing merging guided by various heuristics \citep{Davari_Belilovsky_2023,luo2023lcm,jin2023dataless}, such as resolving interference due to redundant parameter values and sign disagreements \citep{Yadav_Tam_Choshen_etal_2023}, by preserving the important parameters defined via the Fisher Information Matrix \citep{matena2022merging,Tam_Bansal_Raffel_2023}, or by learning the model merging weights with unlabeled test data \citep{Yang_Wang_Shen_Liu_Guo_Wang_Tao_2023}. 
\citet{ortiz2023task} offered more theoretical foundations on the field of \textit{model merging} and identified \textit{weight disentanglement} as the necessary condition for task arithmetic, while showing that performing fine-tuning on a linearized model leads to improved model merging.

\myparagraph{Reducing complexity for foundation models}
The capabilities of foundation models have commanded the development of methods that address their computational and memory requirements. Parameter-efficient fine-tuning (PEFT) \citep{hu2022lora,liu2022few,houlsby2019parameter} approaches heavily reduce the number of trainable parameters, enabling efficient adaptation. Several works \citep{dettmers2023qlora,liang2021pruning} perform  quantization after training and reduce the memory footprint of the model by requiring less bits to represent each parameter. ComPEFT \citep{Yadav_Choshen_Raffel_Bansal_2023} addresses an orthogonal issue, namely communication costs in expert model merging, by compressing fine-tuning residuals via sparsification and quantization. Task Arithmetic can also be viewed from a compression standpoint; multiple functionally diverse models are combined into one, but severe performance degradation is observed. BYOM \cite{jiang2024byom} sparsifies the residuals before merging but the performance heavily depends on the chosen sparsity level. In this paper, we introduce a mechanism that addresses this drop while significantly compressing the multiple initial checkpoints. {Mixture of Experts (MoE) architectures~\citep{shazeer2017moe,riquelme2021scaling}, where different sub-networks specialize in various tasks or input regions, offer an effective approach for handling diverse objectives. However, training such models from scratch can be complex \citep{chen2022towards,fedus2022switch}. This work explores an alternative approach, leveraging the power of pre-trained models and task-specific information localization to create expert models, potentially streamlining MoE development.}
\looseness=-1

\looseness=-1

\section{Task interference causes performance degradation}
\label{sec:preliminaties}

We consider the case of \( \numtasks \) tasks, where  training for each task \( t \) starts from pre-trained model \ptm and fine-tunes on $\calD_t^{\textrm{train}}$
to obtain \( \ttheta_t \). 
Task arithmetic \cite{ilharco2023task} merges the fine-tuned checkpoints by decoupling the contributions of the zero-shot model and operating on the space of residuals or \textit{task vectors} \( \tv{t} =\ttheta_t-\ptm\), generating the \mtv through simple summation: $\tvopt=\sum_{t\in[\numtasks]}\tv{t}$. The final multi-task model corresponds to $\ttheta=\ptm+\alpha\tvopt$, where $\alpha>0$ is a scaling factor tuned on a held-out validation set.\looseness=-1

\looseness=-1
While task arithmetic offers a computationally cheap way to fuse multiple fine-tuned checkpoints, it suffers from significant performance drops compared to the single-task counterparts. Previous works have attributed the performance drop to loss of valuable task-specific information due to parameter interference during merging process \cite{Yadav_Tam_Choshen_etal_2023}. 
To better understand the causes for performance degradation, we make two hypotheses:
\begin{itemize}
    \item \textit{Information erasure}: large amount of information specific to each task is erased when merging the \mtv.
    \item \textit{Task interference}: the task-specific information is preserved in the \mtv, but can not manifest properly due to interference between the tasks.
\end{itemize}

To validate these hypotheses, we start with a controlled experiment where information erasure would not happen. Specifically, for the 8-task vision benchmark proposed by \citet{ilharco2023task}, we randomly select a subset of weights for each task and perform gradient updates only for those parameters. Hence, task vectors {$\{\tv{t}\}_{t=1}^8$} form a partition of the weight space, where all non-overlapping subsets are of equal size. 
By design,  parameter interference \citep{Yadav_Tam_Choshen_etal_2023} is nonexistent   and tasks do not compete for important parameters \citep{matena2022merging,Tam_Bansal_Raffel_2023}. Importantly, all the task-relevant information is preserved inside the \mtv. 
\looseness=-1

The results for this control experiment are presented in \autoref{tab:ortho_fine-tuning}, compared with task arithmetic where the models are fine-tuned in a standard way. Looking at the normalized accuracy, defined in Appendix~\ref{app:experimental details}, we observe that the performance of task arithmetic in the controlled setting deteriorates at the same rate as standard fine-tuning, where the accuracy of the merged model is 2.7\% worse than standard case. 
This suggests that, even when the task-specific knowledge is perfectly preserved inside the \mtv, task arithmetic fails to properly utilize the relevant information to restore the fine-tuned performance. It hints that \interference is the culprit for the performance decline of task arithmetic rather than \textit{weight interference}. 
Specifically, while task-specific parameters remain constant, alterations in other tasks lead to changes in discriminating features for that task, perturbing the mapping from the task's input distribution to output. 
\looseness=-1

\looseness=-1

\begin{table}[t!]
\centering
\caption{Performance comparison between standard and non-overlapping fine-tuning,  averaged over 8 tasks \citep{ilharco2023task}; the lack of \textit{weight interference} and the preservation of all task-specific knowledge in the controlled experiment is not beneficial for task arithmetic.}
\vspace{1em}
\resizebox{\columnwidth}{!}{
\begin{tabular}{@{}cccc@{}}
\toprule
 &  & Abs. acc. & Norm. acc. \\ \midrule
\multirow{2}{*}{Fine-tuning method} & Standard & 92.8 & 100.0 \\
 & Controlled & 89.8 & 100.0 \\ \midrule
\multirow{2}{*}{Task arithmetic accuracy} & Standard & \textbf{71.5} & \textbf{77.0} \\
 & Controlled & 68.8 & 76.6 \\ \bottomrule
\end{tabular}
}
\label{tab:ortho_fine-tuning}
\end{table}

\looseness=-1

\section{\maskname: Localizing task-specific information in \mtv}

In the controlled experiment, the fine-tuning performance can be easily restored by localizing task-specific information in the \mtv with the masks used for fine-tuning.
Now we shift our focus to the general setting of standard fine-tuning and investigate the percentage of information preserved after merging.

We formulate the problem of localization of task-specific knowledge \citep{grafting,dai-etal-2022-knowledge} as extracting relevant weight subsets from the \mtv with binary masks, such that the extracted weights approximate the original task vector $\ttau_t$. The binary mask deactivates irrelevant weights in \mtv while keeping only the task-specific information. Our algorithm, \maskname for \underline{TA}sk \underline{L}oca\underline{L}ization Masks, constructs masks $\mm_t$ 
targeting to construct $\hat{\ttheta}_t$ such that:
\begin{equation}
\label{eq:construct_theta_hat}
\hat{\ttheta_t} = \ptm + \bm{m}_t \circ \tvopt \approx \ttheta_t
\end{equation} 
We minimize the $\ell_1$ distance between the reconstructed $\hat{\ttheta}_t$ and fine-tuned model $\ttheta_t$:
\begin{align}
\bm{m}_{t}^* &= \underset{\bm{m}_t \in \{0,1\}^P}{\operatorname{argmin}} \|\hat{\bm{\theta}}_{t} - \bm{\theta}_{t}\|_1  \\
&= \underset{\bm{m}_t \in \{0,1\}^P}{\operatorname{argmin}} \|\bm{m}_{t} \circ \bm{\tau}_{\text{MTL}} - \bm{\tau}_{t}\|_1 \\
&= \mathds{1} \{ |\bm{\tau}_t| \geq |\bm{\tau}_{\text{MTL}} - \bm{\tau}_t| \}
\label{eq:lambd}
\end{align}
Here $P$ stands for the total number of parameters, 
the detailed derivation are given in Appendix~\ref{app:derivation}. Furthermore, we add a hyper-parameter $\lambda_t$ on the right hand side of Equation \ref{eq:lambd} to tune the amount of information for the mask to extract from \mtv; the smaller $\lambda_t$, the more parameters get selected by $\bm{m}_t$. Finally, we construct the task-specific masks based on: \looseness=-1
\begin{equation}
    \bm{m}_t = \mathds{1} \left\{|\ttau_t| \geq |\tvopt - \ttau_t| \cdot \lambda_t \right\} 
    \label{eq:final masks}
\end{equation}
\looseness=-1
Note that $\lambda_t$ is selected based on the validation accuracy of each task respectively, allowing for the task-specific problems to be solved in parallel and independently.

We validate the efficacy of our mask construction by checking if the original performance in the same 8-task computer vision benchmark, evaluated on a held-out dataset, can be restored. Specifically, we construct the masks for each dataset via \autoref{eq:final masks} for the benchmark proposed by \citet{ilharco2023task}, and evaluate with reconstructed models as in \autoref{eq:construct_theta_hat}. \autoref{fig:mask_percentage}  
confirms that full performance can be retained by simply deactivating irrelevant parameter subsets with binary masks. Thus, it shows that all the information embedded in the original checkpoints is not erased but rather preserved in the \mtv. 
\looseness=-1

\begin{figure}[t]
    \centering
    \includegraphics[width=1\linewidth]{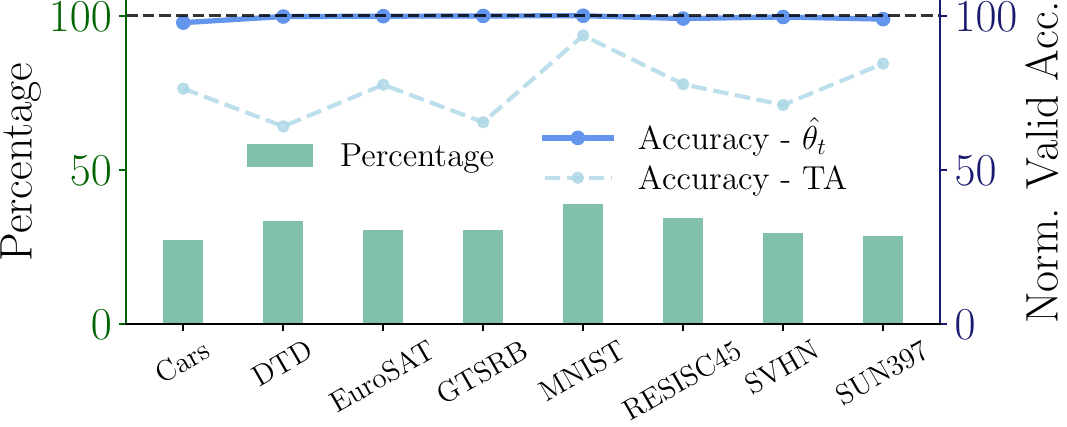}
    \caption{\maskname localizes task-specific information. The bar plot shows the percentage of parameters selected by \maskname, while the blue line shows the normalized validation accuracy achieved by the re-constructed $\hat{\ttheta_t}$ with the selected masks using  \autoref{eq:construct_theta_hat}. The lightblue dashed line shows the task arithmetic baseline where the information is not localized. Our task-specific masks allow the restoration of full performance, showing that all knowledge embedded in the initial fine-tuned checkpoints is preserved post merging. \looseness=-1}
    \label{fig:mask_percentage}
\end{figure}

\section{Applications}
\label{sec:method}

Based on these observations, we present two application scenarios of the masking algorithm for compressing the task vectors and improving model merging methods. 

\subsection{Compressing Task Vectors}
\label{sec:specialist}
Motivated by the previous results, we employ the masks for compressing the fine-tuned checkpoints. Since full performance can be retained by the constructed models using the masks, it allows us to significantly reduce the required storage cost without sacrificing performance.

Specifically, instead of the collection of fine-tuned checkpoints $\{\ttheta_t\}_{t=1}^T$, we can save only the pre-trained model \ptm, the \mtv \tvopt and the individual task-specific {binary} masks $\{\bm{m}_t\}_{t=1}^T$. For evaluation on task $t$, we construct a specialized model by adding to the \pretrained only task-relevant subsets from the \mtv:
\begin{equation}
\hat{\ttheta_t} = \ptm + \bm{m}_t \circ \tvopt
\end{equation}
\looseness=-1
In this way, it allows significant compression while maintaining the majority of performance for fine-tuned models without saving the individual checkpoints. For example, it requires only 13.7\% of storage compared to saving the ViT-L/14 checkpoints for a 20-task benchmark. We provide the details for storage comparison in Appendix~\ref{app:storage}.

\subsection{Improving Model Merging}
\label{sec:generalist}

While the storage of task-specific
masks introduces extra storage cost compared with Task Arithmetic, we present here another application of \maskname, \secondalgoname, which improves over model merging methods without requiring extra storage.

The construction of the task-specific masks $\{\mm_t\}_{t=1}^T$ allows us to investigate the relevance of parameters in the \mtv to each task, where we assume that if a parameter is included in a mask, it is relevant for the associated task.

We find that many parameters in the \mtv are relevant to only a subset of tasks.
Let $P$ be the total number of parameters, we define \textit{mask agreement percentage} as the fraction of weights deemed important by exactly $n$ out of $\numtasks$ tasks:
\looseness=-1
\begin{equation}
    \label{eq:mask agreement}
    \alpha(\{\mm_t\}_{t=1}^T, n)= \frac{1}{P} \left\|\mathds{1} \left\{ \sum_{t\in[T]} \mm_t = n\right\} \right\|_0
\end{equation}
The histogram for the mask agreement percentage is shown in \autoref{fig:selfish weights}, for both the \mtv merged with Task Arithmetic and TIES, while we provide in Appendix~\ref{app:selfish_weights_app} the cases for       more tasks. We observe that there exists a subset of parameters which are used by no task at all, which we term \textit{catastrophic} weights as their existence would only introduce unnecessary interference and hurt the performance of model merging. Furthermore, we also identify there exists a non-negligible fraction of weights used by only one task, which we term \textit{selfish} weights, as their existence only benefits one task whereas causing \interference for all other tasks. We term the rest of weights as \textit{general} weights, as they are relevant to at least two tasks and their importance grows with the number of relevant tasks. Similarly, we term \textit{universal} the weights deemed important for all tasks $(n=T)$.

\begin{figure}[t]
    \centering
    \includegraphics[width=0.95\linewidth]{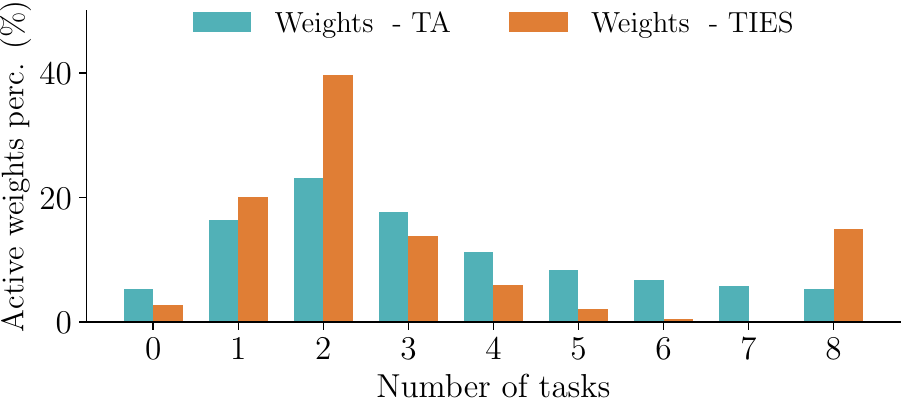}
    \caption{The distribution of mask agreements in the merged vector produced by two model merging methods, Task Arithmetic and TIES. A non-negligent fraction of weights is important exclusively to one task (selfish) while another fraction is irrelevant to all tasks (catastrophic). Our method eliminates both categories to improve model merging. 
    \looseness=-1
    }
    \label{fig:selfish weights}
\end{figure}

\begin{table*}[htbp!]
\centering
\caption{Comparison of model merging (top) and compression (bottom) methods  on three sets of NLP benchmarks with a T5-large model, in terms of accuracy, absolute and normalized (in parentheses), as well as storage cost.}
\vspace{1em}
\resizebox{\textwidth}{!}{
\begin{tabular}{@{}ccccccccccc@{}}
\toprule
\multicolumn{2}{c}{} &  & \multicolumn{2}{c}{7 NLP tasks \cite{Yadav_Tam_Choshen_etal_2023}} &  & \multicolumn{2}{c}{8 QA tasks \cite{zhou2022not}} &  & \multicolumn{2}{c}{All 11 tasks} \\ \cmidrule(lr){4-5} \cmidrule(lr){7-8} \cmidrule(l){10-11} 
\multicolumn{2}{c}{\multirow{-2}{*}{Method}} &  & Acc.(\%) $\uparrow$ & Bits(Gb) $\downarrow$ &  & Acc.(\%) $\uparrow$ & Bits(Gb)  $\downarrow$ &  & Acc.(\%) $\uparrow$ & Bits(Gb) $\downarrow$ \\ \midrule
 & \cellcolor[HTML]{EFEFEF} Zero-shot &   \cellcolor[HTML]{EFEFEF} & \cellcolor[HTML]{EFEFEF}44.9 & \cellcolor[HTML]{EFEFEF}{\color[HTML]{656565} 25.1} & \cellcolor[HTML]{EFEFEF} & \cellcolor[HTML]{EFEFEF}33.1 & \cellcolor[HTML]{EFEFEF}{\color[HTML]{656565} 25.1} & \cellcolor[HTML]{EFEFEF} & \cellcolor[HTML]{EFEFEF}36.9 & \cellcolor[HTML]{EFEFEF}{\color[HTML]{656565} 25.1} \\
 & Weight averaging &  & \foo{60.5}{72.7} & {\color[HTML]{656565} 25.1} &  & \foo{56.4}{69.9} & {\color[HTML]{656565} 25.1} &  & \foo{55.2}{71.6} & {\color[HTML]{656565} 25.1} \\
 & Task arithmetic &  & \foo{71.9}{85.3} & {\color[HTML]{656565} 25.1} &  & \foo{63.8}{79.6} & {\color[HTML]{656565} 25.1} &  & \foo{63.6}{81.8} & {\color[HTML]{656565} 25.1} \\
 & TIES &  & \foo{69.6}{83.5} & {\color[HTML]{656565} 25.1} &  & \foo{62.8}{78.9} & {\color[HTML]{656565} 25.1} &  & \foo{64.0}{82.6} & {\color[HTML]{656565} 25.1} \\
 & \textbf{Consensus TA [ours]} &  & \textbf{\foo{73.5}{87.7}} & {\color[HTML]{656565} 25.1} &  & \foo{68.6}{85.4} & {\color[HTML]{656565} 25.1} &  & \textbf{\foo{67.5}{86.8}} & {\color[HTML]{656565} 25.1} \\
\multirow{-6}{*}{\rotatebox[origin=c]{90}{Merging}} & \textbf{Consensus TIES [ours]} &  & \foo{71.0}{84.2} & {\color[HTML]{656565} 25.1} &  & \textbf{\foo{69.1}{85.8}} & {\color[HTML]{656565} 25.1} &  & \foo{66.8}{85.7} & {\color[HTML]{656565} 25.1} \\ \midrule
 & \cellcolor[HTML]{EFEFEF} Fine-tuned & \cellcolor[HTML]{EFEFEF} & \cellcolor[HTML]{EFEFEF}85.9 & \cellcolor[HTML]{EFEFEF}{\color[HTML]{656565} 169.1} & \cellcolor[HTML]{EFEFEF} & \cellcolor[HTML]{EFEFEF}80.7 & \cellcolor[HTML]{EFEFEF}{\color[HTML]{656565} 193.1} & \cellcolor[HTML]{EFEFEF} & \cellcolor[HTML]{EFEFEF}78.7 & \cellcolor[HTML]{EFEFEF}{\color[HTML]{656565} 265.1} \\
 & Magnitude Pruning &  & \foo{81.6}{93.5} & {\color[HTML]{656565} \textgreater54.3} &  & \foo{70.8}{85.7} & {\color[HTML]{656565} \textgreater55.1} &  & \foo{65.8}{81.4} & {\color[HTML]{656565} \textgreater57.3} \\
 & Magnitude Masking &  & \foo{78.9}{90.7} & {\color[HTML]{656565} 54.3} &  & \foo{72.5}{88.8} & {\color[HTML]{656565} 55.1} &  & \foo{69.8}{87.2} & {\color[HTML]{656565} 57.3} \\
 & \textbf{TALL Mask + TA [ours]} &  & \textbf{\foo{86.8}{102.2}} & {\color[HTML]{656565} 54.3} &  & \foo{79.6}{98.7} & {\color[HTML]{656565} 55.1} &  & \foo{76.5}{96.2} & {\color[HTML]{656565} 57.3} \\
\multirow{-5}{*}{\rotatebox[origin=c]{90}{Compression}} & \textbf{TALL Mask + TIES [ours]} &  & \foo{83.4}{95.4} & {\color[HTML]{656565} 54.3} &  & \textbf{\foo{79.7}{98.8}} & {\color[HTML]{656565} 55.1} &  & \textbf{\foo{77.4}{97.5}} & {\color[HTML]{656565} 57.3} \\ \bottomrule
\end{tabular}
}
\label{tab:nlp_results}
\vspace{1em}
\end{table*}

Based on these observations, we present \secondalgoname, which is targeted to reduce \interference for better model merging. 
{Formally, we form the \textit{consensus mask} for threshold $k\in\{0, \dots,T\}$ as:}
\begin{align}
    \label{eq:consensus}
    \mm_\textrm{consensus} = \mathds{1} \left\{ \sum_{t\in[T]} \mm_t \geq k\right\},
\end{align}
and filter the \mtv through a Hadamard product:
\begin{equation}
    \label{eq:consensus_hadamard}
    \tv{\textrm{consensus}} = \mm_\textrm{consensus}\circ \tvopt,
\end{equation}

where $k$ in \autoref{eq:consensus} is defined as the weight-pruning threshold, e.g., the minimal number of activated masks for preventing the weights from being pruned. By setting $k=2$ in \autoref{eq:consensus} we can eliminate both \textit{catastrophic} weights and \textit{selfish} weights in the \mtv to reduce \interference, keeping only \textit{general} weights that are globally important to at least two tasks. The \textit{threshold} $k$ affects performance and depends on the task number and combination, as well as the underlying model merging method, as \autoref{fig:selfish weights} showcases the different \textit{mask agreement profiles} for Task Arithmetic and TIES. In the following, we use $k=2$, unless specified otherwise.

Finally, we note that both the proposed compression method and \secondalgoname are orthogonal to existing model merging  approaches and can be easily plugged in since they operate on the \mtv \tvopt, e.g., \tvopt can be produced by Task Arithmetic \citep{ilharco2023task}, \ties \citep{Yadav_Tam_Choshen_etal_2023} or other algorithms \citep{matena2022merging,Tam_Bansal_Raffel_2023}. Practitioners can toggle between these two applications depending on the usage scenario. \looseness=-1

\section{Experiments}
\label{sec:experiments}

\looseness=-1

\subsection{Model Merging}

\myparagraph{Baselines} 
We compare \secondalgoname with several train-free model-merging methods, including weight averaging, Task Arithmetic \citep{ilharco2023task}, and \ties \citep{Yadav_Choshen_Raffel_Bansal_2023}. Our method is complementary to all and we opt to validate its efficacy when combined with the latter two. Specifically, we term our methods Consensus Task Arithmetic and Consensus TIES when combined with them respectively.
We include also individually fine-tuned models and the zero-shot model as higher and lower bounds on performance, respectively. We assess the performance based on both the averaged absolute accuracy, and normalized accuracy, defined in detail in Appendix~\ref{app:experimental details}.

\looseness=-1
\myparagraph{Natural Language Processing}
We explore NLP benchmarks following \citet{Yadav_Tam_Choshen_etal_2023,Tam_Bansal_Raffel_2023}. We use a variant of T5-large model \citep{Raffel_Shazeer_Roberts_etal_}, T5-large-LM-Adapt \citep{lester2021power}, and evaluate our method on three sets of benchmarks studied in previous works, a 7-task NLP benchmark \citep{Yadav_Tam_Choshen_etal_2023}, an 8-task benchmark geared towards Question-Answering \citep{zhou2022not}, as well as their union amounting to 11 tasks overall. 
More details about the task composition for each benchmark are provided in \autoref{app:experimental details}. We use the publicly released checkpoints from \citet{Tam_Bansal_Raffel_2023}.
\looseness=-1

\autoref{tab:nlp_results} presents the results for all NLP benchmarks. \secondalgoname consistently improves over both Task Arithmetic and \ties, leading to significant performance gains across settings. For example, in the 8-QA benchmark, our methods respectively improve over Task arithmetic by 4.8\% and \ties by 6.3\% in absolute accuracy, while performance is enhanced by over 2.9\% and 2.8\%, respectively, for the 11-task benchmark. We also conduct experiments with checkpoints finetuned with parameter-efficient methods in (IA)$^3$ \citep{liu2022few};  Appendix ~\ref{appendix:ia3 results} shows that our proposed method again results in significant gains over Task Arithmetic and TIES. 
\looseness=-1

\myparagraph{Computer Vision}
We consider 3 test scenarios, where the number of tasks in each scenario increases gradually from 8 to 14 and 20. The 8-task benchmark coincides with the experimental setup originally introduced by \citet{ilharco2023task} and is expanded to further illuminate the effect of larger number of tasks. The full details for the benchmarks are provided in Appendix~\ref{sec:benchmark contents}. For each test scenario, we assess the efficacy of our method on three CLIP model variants \citep{radford2021learning} with ViT-B/32, ViT-B/16, and ViT-L/14 as visual encoders \citep{dosovitskiy2021an}. All methods use the same checkpoints, fine-tuned with the setting outlined in \citet{ilharco2023task}. 
\looseness=-1

The results in image classification are shown in \autoref{tab:vision_results}. 
Similar to NLP, \secondalgoname provides the best results in 5 out of 6 scenarios, with its superiority becoming increasingly apparent as both the number of tasks and model size grow. This pattern is also noted for the ViT-B/16, as detailed in \autoref{tab:vision_results_vit_b_16} in Appendix~\ref{app:result_vit_b_16}. Our algorithm offers consistent enhancements over Task Arithmetic, while its advantages over \ties are most pronounced in larger models. For example, in the most extensive evaluation involving 20 tasks with a ViT-L/14 encoder, our approach improves on Task Arithmetic and \ties by 4.9\% and 1.1\%, respectively.
\looseness=-1

\begin{table*}[t!]
\centering
\caption{Comparison of  model merging (top) and compression (bottom) methods  across three test scenarios in image classification with different ViT encoders for CLIP models, in terms of accuracy, absolute and normalized (in parentheses), as well as storage cost (in Gb). \looseness=-1}
\vspace{1em}
\renewcommand{\arraystretch}{1.5} 
\resizebox{\textwidth}{!}{
\begin{tabular}{@{}ccccccccccccccccccccc@{}}
\toprule
\multicolumn{2}{c}{} &  & \multicolumn{8}{c}{ViT-B/32} &  &  & \multicolumn{8}{c}{ViT-L/14} \\ \cmidrule(lr){4-11} \cmidrule(l){14-21} 
\multicolumn{2}{c}{} &  & \multicolumn{2}{c}{8 tasks} &  & \multicolumn{2}{c}{14 tasks} &  & \multicolumn{2}{c}{20 tasks} &  &  & \multicolumn{2}{c}{8 tasks} &  & \multicolumn{2}{c}{14 tasks} &  & \multicolumn{2}{c}{20 tasks} \\ \cmidrule(lr){4-5} \cmidrule(lr){7-8} \cmidrule(lr){10-11} \cmidrule(lr){14-15} \cmidrule(lr){17-18} \cmidrule(l){20-21} 
\multicolumn{2}{c}{\multirow{-3}{*}{Method}} &  & Acc.(\%) $\uparrow$ & Bits$\downarrow$ &  & Acc.(\%) $\uparrow$ & Bits$\downarrow$ &  & Acc.(\%) $\uparrow$ & Bits$\downarrow$ &  &  & Acc.(\%) $\uparrow$ & Bits$\downarrow$ &  & Acc.(\%) $\uparrow$ & Bits$\downarrow$ &  & Acc.(\%) $\uparrow$ & Bits$\downarrow$ \\ \midrule
\rowcolor[HTML]{EFEFEF} 
\cellcolor[HTML]{FFFFFF} & Zeroshot &  & 48.4 & {\color[HTML]{656565} 3.6} &  & 57.3 & {\color[HTML]{656565} 3.6} &  & 56.1 & {\color[HTML]{656565} 3.6} &  &  & 64.4 & {\color[HTML]{656565} 11.0} &  & 68.0 & {\color[HTML]{656565} 11.0} &  & 65.1 & {\color[HTML]{656565} 11.0} \\

\cellcolor[HTML]{FFFFFF} & Weight averaging &  & \foo{66.5}{72.3} & {\color[HTML]{656565} 3.6} &  & \foo{64.4}{71.2} & {\color[HTML]{656565} 3.6} &  & \foo{61.1}{67.5} & {\color[HTML]{656565} 3.6} &  &  & \foo{79.4}{83.0} & {\color[HTML]{656565} 11.0} &  & \foo{76.6}{81.0} & {\color[HTML]{656565} 11.0} &  & \foo{71.5}{75.5} & {\color[HTML]{656565} 11.0} \\

\cellcolor[HTML]{FFFFFF} & Task arithmetic &  & \foo{70.8}{76.5} & {\color[HTML]{656565} 3.6} &  & \foo{65.4}{72.2} & {\color[HTML]{656565} 3.6} &  & \foo{60.6}{66.8} & {\color[HTML]{656565} 3.6} &  &  & \foo{84.8}{88.5} & {\color[HTML]{656565} 11.0} &  & \foo{79.3}{83.8} & {\color[HTML]{656565} 11.0} &  & \foo{74.0}{78.0} & {\color[HTML]{656565} 11.0} \\
\cellcolor[HTML]{FFFFFF} & TIES &  & \textbf{\foo{75.1}{81.0}} & {\color[HTML]{656565} 3.6} &  & \foo{68.0}{74.8} & {\color[HTML]{656565} 3.6} &  & \foo{63.4}{69.9} & {\color[HTML]{656565} 3.6} &  &  & \textbf{\foo{86.9}{90.7}} & {\color[HTML]{656565} 11.0} &  & \foo{79.5}{84.1} & {\color[HTML]{656565} 11.0} &  & \foo{75.7}{79.8} & {\color[HTML]{656565} 11.0} \\

\cellcolor[HTML]{FFFFFF} & \textbf{Consensus TA [ours]} &  & \foo{75.0}{80.8} & {\color[HTML]{656565} 3.6} &  & \textbf{\foo{70.4}{77.4}} & {\color[HTML]{656565} 3.6} &  & \textbf{\foo{65.4}{72.0}} & {\color[HTML]{656565} 3.6} &  &  & \foo{86.2}{89.9} & {\color[HTML]{656565} 11.0} &  & \textbf{\foo{82.2}{86.9}} & {\color[HTML]{656565} 11.0} &  & \textbf{\foo{78.9}{83.2}} & {\color[HTML]{656565} 11.0} \\

\multirow{-6}{*}{\cellcolor[HTML]{FFFFFF}\rotatebox[origin=c]{90}{Merging}} & \textbf{Consensus TIES [ours]} &  & \foo{74.8}{80.6} & {\color[HTML]{656565} 3.6} &  & \foo{67.7}{74.5} & {\color[HTML]{656565} 3.6} &  & \foo{63.2}{69.6} & {\color[HTML]{656565} 3.6} &  &  & \textbf{\foo{86.9}{90.7}} & {\color[HTML]{656565} 11.0} &  & \foo{81.5}{86.1} & {\color[HTML]{656565} 11.0} &  & \foo{76.8}{80.9} & {\color[HTML]{656565} 11.0} \\ \midrule
\rowcolor[HTML]{EFEFEF} 
\cellcolor[HTML]{FFFFFF}{\color[HTML]{333333} } & \cellcolor[HTML]{EFEFEF}{\color[HTML]{333333} Fine-tuned} & \cellcolor[HTML]{EFEFEF} & \cellcolor[HTML]{EFEFEF}92.8 & {\color[HTML]{656565} 23.3} &  & 90.9 & {\color[HTML]{656565} 40.2} &  & 91.3 & {\color[HTML]{656565} 57.0} &  &  & 95.8 & {\color[HTML]{656565} 79.1} &  & 94.3 & {\color[HTML]{656565} 137.4} &  & 94.7 & {\color[HTML]{656565} 195.8} \\
\cellcolor[HTML]{FFFFFF}{\color[HTML]{333333} } & Magnitude Pruning &  & \foo{91.3}{98.4} & {\color[HTML]{656565} \textgreater7.1} &  & \foo{85.3}{93.7} & {\color[HTML]{656565} \textgreater7.7} &  & \foo{83.4}{91.2} & {\color[HTML]{656565} \textgreater8.2} &  &  & \foo{95.4}{99.6} & {\color[HTML]{656565} \textgreater23.1} &  & \foo{91.8}{97.2} & {\color[HTML]{656565} \textgreater24.9} &  & \foo{91.2}{96.1} & {\color[HTML]{656565} \textgreater26.8} \\
\cellcolor[HTML]{FFFFFF}{\color[HTML]{333333} } & Magnitude Masking &  & \foo{86.8}{93.3} & {\color[HTML]{656565} 7.1} &  & \foo{80.7}{88.4} & {\color[HTML]{656565} 7.7} &  & \foo{75.3}{82.1} & {\color[HTML]{656565} 8.2} &  &  & \foo{94.6}{98.7} & {\color[HTML]{656565} 23.1} &  & \foo{91.6}{97.0} & {\color[HTML]{656565} 24.9} &  & \foo{91.6}{96.5} & {\color[HTML]{656565} 26.8} \\
\cellcolor[HTML]{FFFFFF}{\color[HTML]{333333} } & \textbf{TALL Mask + TA [ours]} &  & \foo{92.6}{99.7} & {\color[HTML]{656565} 7.1} &  & \foo{90.1}{99.1} & {\color[HTML]{656565} 7.7} &  & \foo{90.6}{99.2} & {\color[HTML]{656565} 8.2} &  &  & \foo{95.7}{99.9} & {\color[HTML]{656565} 23.1} &  & \foo{93.1}{98.8} & {\color[HTML]{656565} 24.9} &  & \foo{93.7}{98.9} & {\color[HTML]{656565} 26.8} \\
\multirow{-5}{*}{\cellcolor[HTML]{FFFFFF}{\color[HTML]{333333} \rotatebox[origin=c]{90}{Compression}}} & \textbf{TALL Mask + TIES [ours]} &  & \textbf{\foo{93.0}{100.3}} & {\color[HTML]{656565} 7.1} &  & \textbf{\foo{90.9}{100.0}} & {\color[HTML]{656565} 7.7} &  & \textbf{\foo{91.1}{99.7}} & {\color[HTML]{656565} 8.2} &  &  & \textbf{\foo{95.9}{100.1}} & {\color[HTML]{656565} 23.1} &  & \textbf{\foo{93.4}{99.0}} & {\color[HTML]{656565} 24.9} &  & \textbf{\foo{93.9}{99.1}} & {\color[HTML]{656565} 26.8} \\ \bottomrule
\end{tabular}
}
\label{tab:vision_results}
\end{table*}

\looseness=-1
\subsection{Compression} 

We adopt the compression technique discussed in \autoref{sec:specialist} to compress the individual checkpoints, and use the prefix `\textit{TALL Mask +}' when combined with different model merging methods. 

\myparagraph{Baselines}
We compare against methods on the same level of storage as our proposed solution. We first consider Magnitude Masking, where per-task masks are constructed by replacing our procedure in \autoref{eq:lambd} by  keeping the top $k\%$ of the parameters. After experimental validation, we found that $k=10$ works well in general and therefore use it throughout. We also consider unstructured Magnitude Pruning of the individual task vectors, where we set the level of pruning so that the storage cost of one pre-trained model and $T$ pruned task vectors is equal to ours. Note that, in practice, Magnitude Pruning takes more storage than our method as we do not consider the storage cost for the positions of the parameters.
\looseness=-1

\begin{figure*}[t!]
    \centering
    \includegraphics[width=\linewidth]{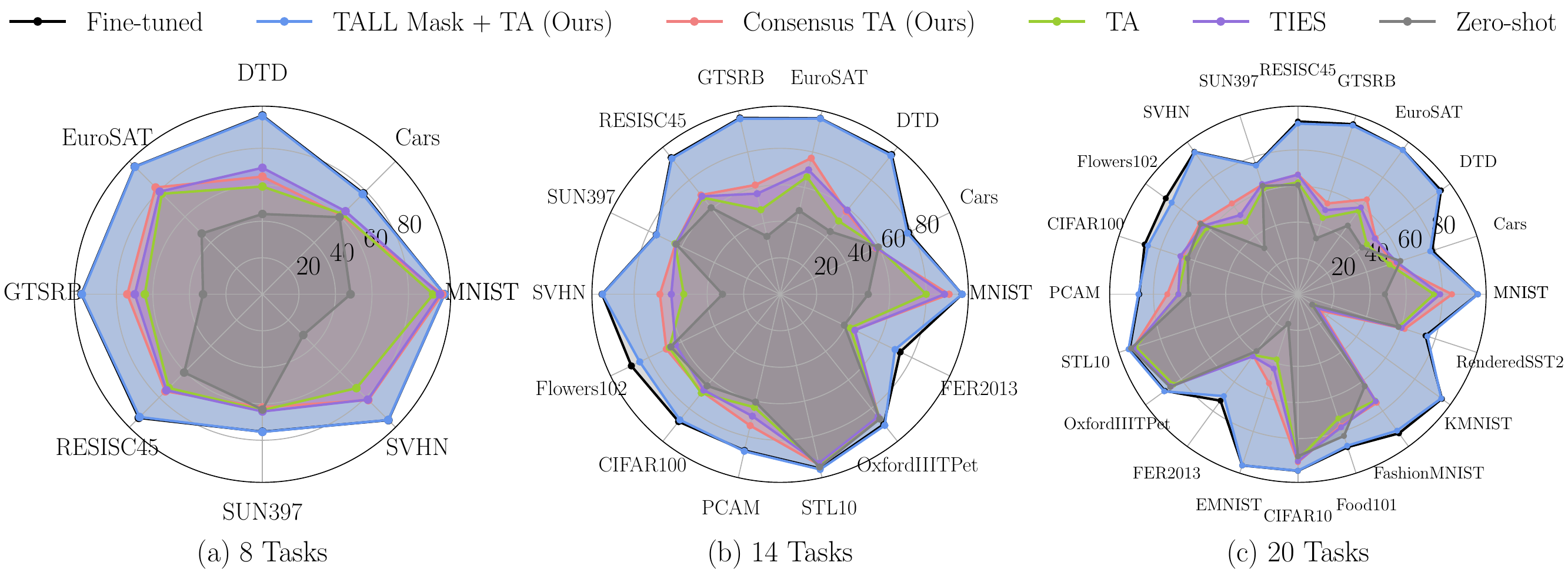}
    \vspace*{-15pt}
    \caption{Comparison of absolute accuracy (\%) of individual tasks for the computer vision benchmarks and ViT-B/32. Results for ViT-B/16 and ViT-L/14 are provided in the appendix. 
    Our \secondalgoname shows higher performance compared to model merging baselines, especially for the settings with more tasks. 
    Our compression algorithm consistently matches the performance of the individual fine-tuned models at a fraction of the memory, while model merging techniques are not robust to the increase of tasks. }
    \vspace{1.2em}
    \label{fig:radar_plot_vitb32}
\end{figure*}

\begin{figure*}[ht]
    \centering
    \includegraphics[width=\linewidth]{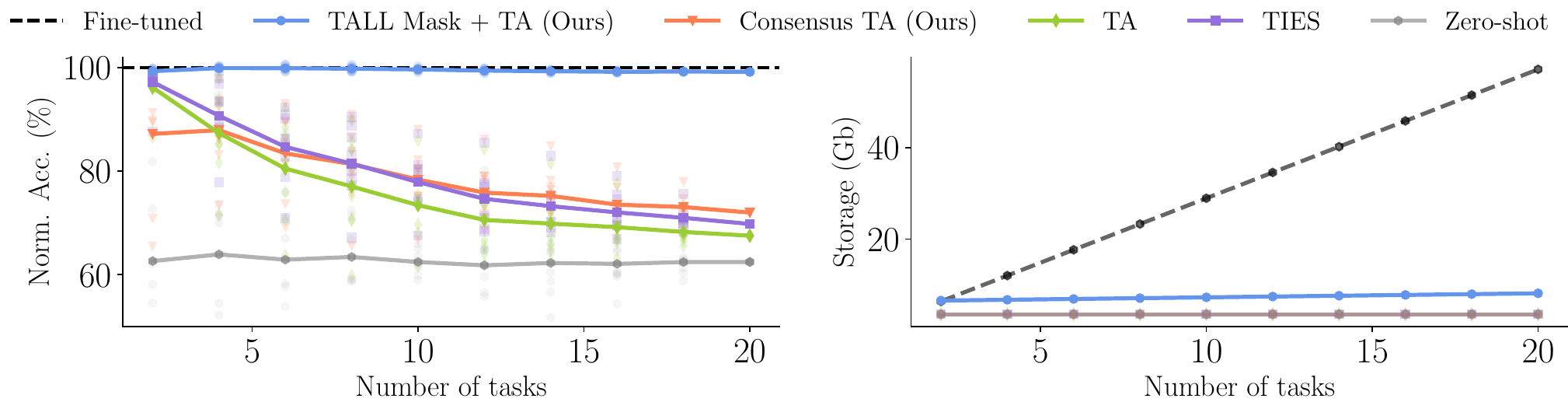}
    \vspace*{-15pt}
    \caption{Averaged normalized accuracy \emph{vs}. number of tasks for computer vision benchmarks. Our proposed specialist algorithm maintains initial performance regardless of task combination and heavily compresses the fine-tuned checkpoints.    }
    \label{fig:num_tasks vs acc and storage}
\end{figure*}

\myparagraph{Natural Language Processing}
Following the same experimental setting for merging, the results are presented in the bottom half of \autoref{tab:nlp_results}. Our compression scheme effectively retains  performance while requiring much less storage cost than storing individual fine-tuned models. In contrast, magnitude-based solutions suffer from severe performance losses, especially as the number of tasks increases. 
For example, for the 7-task benchmark, our method keeps all the performance from the fine-tuned models with even over 100\% normalized accuracy, while requiring less than 1/3 storage cost than storing the checkpoints. The advantage on storage compression becomes more pronounced with larger number of tasks. For example, for the 11-task benchmark, both our methods require only around 1/5 of the storage of individual checkpoints, while keeping at least 96.2\% of the performance. 
By keeping the same level of storage, separately pruned models preserve lower number of parameters and the cost in performance mirrors this lack of task-specific features; for the 11-task benchmark only 81.4\% of performance is retained compared to our performance of $97.5\%$. In contrast, our method takes advantage of the redundancy in the task vector to compress effectively with minimal performance losses.

\myparagraph{Computer Vision}
The bottom half of \autoref{tab:vision_results} presents the results for compression in vision. While we observe that both Magnitude Pruning and Magnitude Masking show a clear performance degradation with increasing number of tasks, our proposed methods deliver almost full performance in all test scenarios. 

Specifically, TALL Mask + Task Arithmetic achieves around 99\% normalized accuracy on all cases, where there is almost no performance degradation with increasing number of tasks. TALL Mask + TIES performs even better, with its normalized accuracy being over 99\% for all the test cases. For ViT-B/32, it achieves around 100\% for all test scenarios without any loss of performance. For ViT-L/14, TALL Mask + TIES still performs exceptionally, with its normalized accuracy being around 100\% for 8 tasks and over 99\% for 14 and 20 tasks.
These results show that our method is able to capture the crucial task-specific information buried in the merged vector and is not bound by model scale or number of tasks. 
\looseness=-1

In terms of storage, TALL Mask + Task Arithmetic requires much less storage cost compared to storing the individual fine-tuned models, where the storage saving is more pronounced with a larger number of tasks. Importantly, our compression scheme with \maskname, provides an efficient trade-off between performance and storage compared to the model-merging methods. For example, using TALL Mask + Task Arithmetic on 20 tasks with ViT-B/32 takes 8.2 Gb storage while achieving 90.6\% absolute accuracy, whereas using TIES on the same 20 tasks with ViT-L/14 takes 11.0 Gb but delivers an absolute accuracy of merely 75.7\%. Overall, our method delivers a desirable trade-off in the Pareto Front of performance retention vs total storage cost.  
\looseness=-1

\subsection{Individual-task performance}

We now shift our focus from statistics over all tasks and present the performance on individual tasks. For vision settings, we compare the performance of TALL Mask + Task Arithmetic and Consensus Task Arithmetic with baselines methods, and plot in \autoref{fig:radar_plot_vitb32} the individual accuracies on three benchmarks with ViT-B/32, while we provide the vision results with two other models as well as the results in NLP settings in Appendix~\ref{app:individual_accuracies}. We observe that TALL Mask + Task Arithmetic consistently delivers performance the same level as individual fine-tuned models, across datasets and total number of tasks. 
\looseness=-1

On the other hand, the expansion of the considered tasks results in significant performance drop for model merging methods, where for some datasets the performance is even reset back to zero-shot model. Yet, we observe that Consensus Task Arithmetic suffers the least from increasing number of tasks and shines among the model merging methods, especially in the 20-task setting, where it outperforms the other model merging methods in almost all the datasets.

\looseness=-1

\subsection{Performance with varying task combinations}

We perform a systematic study on the effect of different task combinations on our methods and model-merging baselines, and present both the normalized accuracy and the storage cost with respect to different numbers of tasks in \autoref{fig:num_tasks vs acc and storage}. 
Due to the vast array of $2^{20}$ potential combinations, the performance is averaged on 8 carefully selected representative sample task combinations for each number of tasks;  the details for selection process are given in \autoref{app:sample_selection}. 
Our observations indicate that our TALL Mask + Task Arithmetic consistently matches the performance of individually fine-tuned models across all combinations, regardless of the total task count. By comparison, model merging methods show a stark degradation in performance, especially given a large number of tasks where their performance degrade gradually to nearly zero-shot performance. Nonetheless, Consensus Task Arithmetic outperforms the two other model merging methods in general, while the performance gain gradually becomes apparent with larger number of tasks. 
\looseness=-1

While the model merging methods have constant storage cost, their applicability is undermined by low performance. Conversely, maintaining individual models is excessively prohibitive in terms of storage but guarantees strong performance.
Localizing the task-specific parameters with our proposed masks offers a favorable trade-off between performance and storage. Crucially, we deliver consistent performance with near 100\% normalized accuracy across various task combinations and numbers of tasks, while heavily compressing the required storage cost, as we see in the right plot of \autoref{fig:num_tasks vs acc and storage}. It presents a more viable solution in scenarios where balancing performance with storage efficiency is essential.
\looseness=-1

\begin{figure}[t]
    \centering
    \includegraphics[width=1\linewidth]{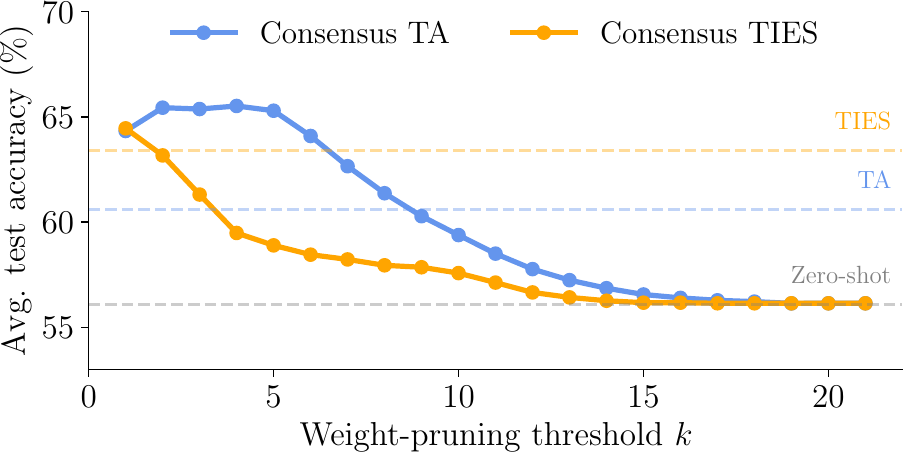}
    \vspace*{-15pt}
    \caption{Performance of Consensus Merging with varying weight-pruning threshold $k$, on the 20-task image classification benchmark with ViT-B/32.} %
    \label{fig:acc_vary_k_20task}
\end{figure}

\subsection{Effect of weight-pruning threshold}
\label{sec:k ablation}

The weight-pruning threshold $k$ is used to form the consensus mask in \autoref{eq:consensus} and, determines least number of activated tasks to prevent weights from being pruned. 
Extending the number of tasks, modifying the task combination and the model merging method itself affect the \textit{mask agreement profile}, defined in \autoref{eq:mask agreement}, and consequently the optimal pruning-threshold and the overall performance. 
We present the performance of Consensus Merging with ViT-B/32 on various image classification benchmarks in \autoref{fig:acc_vary_k_20task} for the  20-task benchmark and \autoref{fig:acc_vary_k_8_14task} in the appendix for the 8 and 14-task benchmarks when $k$ gradually increases from 1 to over the total number of tasks. We observe from the figure that while the optimal performance of Consensus TA is usually achieved by setting $k=2$, i.e., removing both \textit{catastrophic} and \textit{selfish} weights, Consensus TIES achieves its optimal performance by setting $k=1$, i.e., removing only catastrophic weights. We present also the results for removing only catastrophic weights in Appendix~\ref{app:remove_catastrophic_weights} for all test scenarios with Computer Vision, where we observe the performance of Consensus TIES consistently outperform TIES.
\looseness=-1

The difference in optimal thresholds $k$ between Task Arithmetic and TIES originates in their profiles, as shown in \autoref{fig:selfish weights} and \autoref{fig:selfish weights app}; the pruning and sign resolution mechanisms of TIES shift the distribution towards increased \textit{universal} and \textit{selfish} weights. Removing the latter altogether results in a significant reduction of salient weights. Studying how different merging strategies affect the weight profiles remains an interesting future direction.
\looseness=-1

\section{Conclusion} 
\label{sec:conclusion}

In this study, we introduced \maskname, a method with dual practical applications: it effectively addresses the significant issue of compressing foundation models and facilitates model merging. We identify the reason for performance degradation of model merging methods is not due to information erasure during merging process, but because of task interference during evaluation. Our research demonstrates that the multi-task vector retains crucial task-specific information, which \maskname localizes and extracts through the use of task-specific binary masks, thereby enabling the recovery of original fine-tuned performance levels. We utilize this observation to compress a collection of fine-tuned checkpoints into storing only the zero-shot model, the information-rich merged vector and the specified masks. Further, our examination of mask statistics uncovered weights harmful to model merging. We thus proposed their elimination and achieved state-of-the-art performance in common model merging benchmarks. Our proposed solution not only enhances the utility of foundation models but also makes them more accessible and sustainable, paving the way for broader applications and advancements in the field.
\looseness=-1

\section*{Impact Statements}

This paper presents work whose goal is to advance the field of Machine Learning. There are many potential societal consequences of our work, none which we feel must be specifically highlighted here.

\section*{Acknowledgments}

We thank Alessandro Favero, Thibault Sejourn\'e and the anonymous reviewers for helpful feedback and comments.

\bibliography{main}
\bibliographystyle{icml2024}

\newpage
\appendix
\onecolumn

\section{Experimental Details}
\label{app:experimental details}

All our experiments were performed using the same hardware consisting of four V100 NVIDIA GPUs with 32GB of memory each.

\subsection{Fine-tuning}

\textbf{Fine-tuning}: For all fine-tuning experiments, we stick to the training procedure outlined in \citet{ilharco2023task}. Specifically, we fine-tune the same pre-trained CLIP checkpoint obtained from the \textit{openclip} repository \cite{openclip}. We fine-tune for 2,000 iterations, using a batch size of 128, a learning rate of $1e^{-5}$, and a cosine annealing learning rate schedule with 200 warm-up steps, along with the AdamW optimizer. Following \citet{ilharco2023task} and \citet{ortiz2023task}, we freeze the weights of the classification layer during fine-tuning process. 

\textbf{Normalized Accuracy} To account for the task difficulties, we provide the normalized accuracies as well as the absolute accuracies in our results. Specifically, the normalization
is performed with respect to the accuracy achieved by the individual fine-tuned models:

\begin{equation}
    \text{Normalized Accuracy} =  \frac{1}{T} \sum_{t=1}^{T} \frac{\underset{x\sim\mu_t}{\text{acc}}\left[f_\text{merged}(x)\right]}{\underset{x\sim\mu_t}{\text{acc}}\left[f_\text{fine-tuned}(x)\right]}
\end{equation}

Note that the normalized accuracy depends on the fine-tuning methods as well as the merging methods.

\subsection{Hyper-parameter tuning} 

\paragraph{Mask sparsity factor $\lambda$} For constructing task-specific masks, we tune the hyper-parameter $\lambda$ for each task over \{0.2, 0.3, 0.4, 0.5, 0.6\}. The best $\lambda$ for each task is selected based the validation performance on each individual tasks. 

\paragraph{Task vector scaling factor $\alpha$} Following \citet{ilharco2023task}, we use a single scaling factor $\alpha$ to scale the multi-task vector for the {model merging} methods in \autoref{tab:nlp_results} and \autoref{tab:vision_results}. The scaling factor is tuned over a range of \{0.0, 0.1, ..., 0.9, 1.0\}, selected based on the performance on the validation set averaged on all tasks. 

\subsection{Benchmark task contents}
\label{sec:benchmark contents}

\subsubsection{Computer vision}

The \textbf{8-task scenario} takes into account the 8 tasks studied in \citet{radford2021learning}, \citet{ilharco2023task}, including
\begin{inparaenum}
    \item Cars \citep{krause20133d},
    \item DTD \citep{dtd},
    \item EuroSAT \citep{HelberBDB19},
    \item GTSRB \citep{gtsrb},
    \item MNIST \citep{lecun1998mnist},
    \item RESISC45 \citep{cheng2017remote},
    \item SUN397 \citep{xiao2016sun},
    \item SVHN \citep{netzer2011reading}.
\end{inparaenum} 

The \textbf{14-task scenario} adds to the 8 tasks mentioned above the following tasks: 
\begin{inparaenum}
    \setcounter{enumi}{8}
    \item CIFAR100 \citep{krizhevsky2009learning},
    \item STL10 \citep{stl10},
    \item Flowers102 \citep{nilsback2008automated},
    \item OxfordIIITPet \citep{parkhi2012cats},
    \item PCAM \citep{veeling2018rotation},
    \item FER2013 \citep{goodfellow2013challenges}.
\end{inparaenum}

The \textbf{20-task scenario} adds to the 14 tasks mentioned above the following tasks: 
\begin{inparaenum}
    \setcounter{enumi}{14}
    \item EMNIST \citep{cohen2017emnist},
    \item CIFAR10 \citep{krizhevsky2009learning},
    \item Food101 \citep{bossard14},
    \item FashionMNIST \citep{xiao2017fashion},
    \item RenderedSST2 \citep{socher2013recursive,radford2019language}
    \item KMNIST \citep{clanuwat2018deep},
\end{inparaenum}

For the benchmarks beyond 8 tasks, we use available datasets from \texttt{torchvision} library. For the 14-task scenario, we aim for diversisty in the tasks as much as possible. After removing the MNIST-variants in the 20-task benchmark, we rank the tasks based on the performance comparison for zero-shot CLIP v.s. linear probe ResNet-50 \cite{radford2019language} with ascending order and select the top 6 tasks.

\subsubsection{Natural language Processing}

\textbf{7 NLP Tasks} This benchmark is studied in \citet{Yadav_Tam_Choshen_etal_2023}
and contains the following datasets
\begin{inparaenum}
    \item QASC \citep{allenai:qasc},
    \item QuaRTz \citep{tafjord-etal-2019-quartz},
    \item PAWS \citep{zhang-etal-2019-paws},
    \item Story Cloze \citep{sharma-etal-2018-tackling},
    \item WikiQA \citep{yang-etal-2015-wikiqa},
    \item Winogrande \citep{WinoGrande} and
    \item WSC \citep{levesque2012winograd}
\end{inparaenum}

\textbf{8 QA Tasks} Following \citet{Tam_Bansal_Raffel_2023}, we evaluate on another benchmark,
containing the following tasks 
\begin{inparaenum}
    \item CosmosQA \cite{huang2019cosmos}, 
    \item QASC \citep{allenai:qasc}, 
    \item QuAIL \cite{rogers2020getting}, 
    \item QuaRTz \citep{tafjord-etal-2019-quartz}), 
    \item PAWS \citep{zhang-etal-2019-paws}.
    \item ROPES \cite{lin2019reasoning},
    \item SocialIQA \cite{sap2019socialiqa},
    \item Wiki QA \citep{yang-etal-2015-wikiqa}.
\end{inparaenum}

\section{Derivation of \autoref{eq:lambd}}
\label{app:derivation}

This short section shows the derivation of the mask criterion in more detail:
\begin{align}
\bm{m}_{t}^* &= \underset{\bm{m}_t \in \{0,1\}^P}{\operatorname{argmin}} (\|\hat{\bm{\theta}}_{t} - \bm{\theta}_{t}\|_1)  \\
&= \underset{\bm{m}_t \in \{0,1\}^P}{\operatorname{argmin}} (\|\bm{m}_{t} \circ \bm{\tau}_{\text{MTL}} - \bm{\tau}_{t}\|_1) \\
&= \underset{\bm{m}_t \in \{0,1\}^P}{\operatorname{argmin}} \sum_{n=1}^P |m^{(n)}_{t} \cdot \tau^{(n)}_{\text{MTL}} - \tau^{(n)}_{t} |  \label{deriv_4} \\
\implies m^{(n)*}_{t} &= \underset{m_t^{(n)} \in \{0,1\}}{\operatorname{argmin}} |m^{(n)}_{t} \cdot \tau^{(n)}_{\text{MTL}} - \tau^{(n)}_{t} | \label{deriv_5} \\
&= \begin{cases}
1 & \text{if} \ |\tau^{(n)}_{t}| \geq |\tau^{(n)}_{\text{MTL}} - \tau^{(n)}_{t}| \\
0 & \textrm{otherwise}
\end{cases}   \\
&= \mathds{1} \{ |\tau^{(n)}_{t}| \geq |\tau^{(n)}_{\text{MTL}} - \tau^{(n)}_{t}| \}  
\end{align}

From \autoref{deriv_4} to \autoref{deriv_5} we use the independence of each sub-problem. Aggregating over all sub-problems yields that the optimal mask is given by:

\begin{equation}    
    \mathbf{m}_t^* = \mathds{1} \{ |\bm{\tau}_t| \geq |\bm{\tau}_{\text{MTL}} - \bm{\tau}_t| \}
\label{eq:lambd2}
\end{equation}

\section{Storage cost calculation}
\label{app:storage}

{This section show the calculation of} the storage cost for each method in \autoref{tab:nlp_results} and \autoref{tab:vision_results}. Let $T$ be the number of tasks, $P$ be the number of all parameters, $P'$ be the number of trainable parameters in the model, and $F$ be the number of frozen parameters in the model. Assuming one float parameter takes 32 bits, for each method, their respective storage cost for $T$ tasks is calculated as:

\begin{itemize}
    \item Fine-tuned models: $32(TP'+F)$. $32TP'$ is for storing $T$ trainable parameters and $32F$ is for storing frozen parameters.
    \item Task arithmetic: $32P$; Stores a single model.
    \item Ties-merging: $32P$; Stores a single model.
    \item Consensus Task arithmetic: $32P$; Stores a single model.
    \item Consensus Ties: $32P$; Stores a single model.
    \item Zero-shot: $32P$; Stores a single model.
    \item Magnitude Masking: $(64+T)P' + 32F$; $64P'+32F$ is for storing zero-shot model and multi-task vector, while $TP'$ is for storing $T$ binary masks.
    \item Magnitude Pruning: $>(64+T)P' + 32F$; For Magnitude Pruning, we need to store a single model, as well as the sparsified task vectors for each task. We calculate the respective sparsity such that the total storage cost will be higher than $(64+T)P' + 32F$.
    \item TALL Mask + Task arithmetic: $(64+T)P' + 32F$; $64P'+32F$ is for storing zeroshot model and multi-task vector, while $TP$' is for storing T binary masks.
    \item TALL Mask + Ties-merging: $(64+T)P' + 32F$; $64P'+32F $is for storing zeroshot model and multi-task vector, while $TP'$ is for storing T binary masks.
    
\end{itemize}

\section{Additional results}

\subsection{Performance for removing only catastrophic weights}
\label{app:remove_catastrophic_weights}
For Consensus Merging, we propose to remove both catastrophic weights and selfish weights, as their contribution to the multi-task performance is limited. In this section, we study the performance of Consensus Merging when removing only catastrophic weights, i.e., weights beneficial for none of the tasks. The results for the experiments in image classification are presented in Table~\ref{tab:acc_k1_cv_1} with ViT-B/32 and ViT-L/14, and Table~\ref{tab:acc_k2_cv_1} with ViT-B/16.

\begin{table}[ht]
\centering
\caption{Performance of Consensus when removing only catastrophic weights with ViT-B/32 and ViT-L/14 on image classification benchmarks.}
\resizebox{1.02\textwidth}{!}{
\begin{tabular}{@{}clcclclcllclclc@{}}
\toprule
 &  & \multicolumn{5}{c}{ViT-B/32} &  &  & \multicolumn{5}{c}{ViT-L/14} \\ \cmidrule{3-7} \cmidrule{10-14}
 &  & 8 tasks &  & 14 tasks &  & 20 tasks &  &  & 8 tasks &  & 14 tasks &  & 20 tasks \\ \midrule
Task arithmetic &  & \foo{70.8}{76.5} &  & \foo{65.4}{72.2} &  & \foo{60.6}{66.8} &  &  & \foo{84.8}{88.5} &  & \foo{79.3}{83.8} &  & \foo{74.0}{78.0} \\
Consensus TA $(k=1)$ &  & \foo{73.8}{79.4} &  & \foo{69.0}{75.8} &  & \foo{64.5}{71.0} &  &  & \foo{85.5}{89.2} &  & \foo{81.4}{86.1} &  & \foo{77.7}{81.9} \\
Consensus TA $(k=2)$ &  & \foo{75.0}{80.7} &  & \textbf{\foo{70.4}{77.4}} &  & \textbf{\foo{65.4}{72.0}} &  &  & \foo{86.2}{89.9} &  & \textbf{\foo{82.3}{86.9}} &  & \textbf{\foo{78.9}{83.2}} \\
TIES &  & \foo{75.2}{81.1} &  & \foo{68.0}{74.9} &  & \foo{63.3}{69.8} &  &  & \foo{86.9}{90.7} &  & \foo{79.5}{84.1} &  & \foo{75.7}{79.8} \\
Consensus TIES $(k=1)$ &  & \textbf{\foo{75.9}{81.0}} &  & \foo{69.0}{75.9} &  & \foo{64.5}{71.0} &  &  & \textbf{\foo{87.5}{91.3}} &  & \foo{81.6}{86.3} &  & \foo{78.0}{82.1} \\
Consensus TIES $(k=2)$ &  & \foo{74.8}{80.6} &  & \foo{67.7}{74.5} &  & \foo{63.2}{69.6} &  &  & \foo{86.9}{90.7} &  & \foo{81.5}{86.1} &  & \foo{76.8}{80.9} \\ \bottomrule
\end{tabular}
}
\label{tab:acc_k1_cv_1}
\end{table}

\begin{table}[h]
\centering
\caption{Performance of Consensus when removing only catastrophic weights with ViT-B/16 on image classification benchmarks.}
\begin{tabular}{@{}cccccccc@{}}
\toprule
 &  &  & \multicolumn{5}{c}{ViT-B/16} \\ \midrule
 &  &  & 8 tasks &  & 14 tasks &  & 20 tasks \\ \midrule
Task arithmetic &  &  & \foo{76.2}{80.6} &  & \foo{70.5}{75.8} &  & \foo{65.7}{70.9} \\
Consensus TA $(k=1)$ &  &  & \foo{78.0}{82.4} &  & \foo{72.9}{78.4} &  & \foo{68.9}{74.0} \\
Consensus TA $(k=2)$ &  &  & \foo{79.2}{83.6} &  & \foo{74.3}{79.8} &  & \textbf{\foo{70.0}{75.2}} \\
TIES &  &  & \textbf{\foo{79.7}{84.3}} &  & \foo{73.2}{78.7} &  & \foo{68.1}{73.3} \\
Consensus TIES $(k=1)$ &  &  & \foo{79.4}{83.8} &  & \textbf{\foo{74.4}{80.0}} &  & \foo{69.9}{75.1} \\
Consensus TIES $(k=2)$ &  &  & \foo{79.4}{83.9} &  & \foo{74.1}{79.5} &  & \foo{68.7}{73.9} \\ \bottomrule
\end{tabular}
\label{tab:acc_k2_cv_1}
\end{table}

\subsection{Effect of weight-pruning threshold for 8 and 14 tasks}

We plot the performance for Consensus Merging with varying weight-pruning threshold $k$, on the 8-task and 14-task benchmarks in image classification with ViT-B/32, in Figure~\ref{fig:acc_vary_k_8_14task}. Similar to the observation from Figure\ref{fig:acc_vary_k_20task}, the performance differs from one method to another (TIES and task arithmetic in our case). While the optimal performance of Consensus TA is achieved by removing both catastrophic and selfish weights (setting the threshold to 2), for Consensus TIES it delivers the best results by removing only the catastrophic weights (setting the threshold to 1). Gradually the performance of Consensus Merging reduces to zero-shot as more and more weights get removed.

\begin{figure*}[t]
    \centering
    \includegraphics[width=1\linewidth]{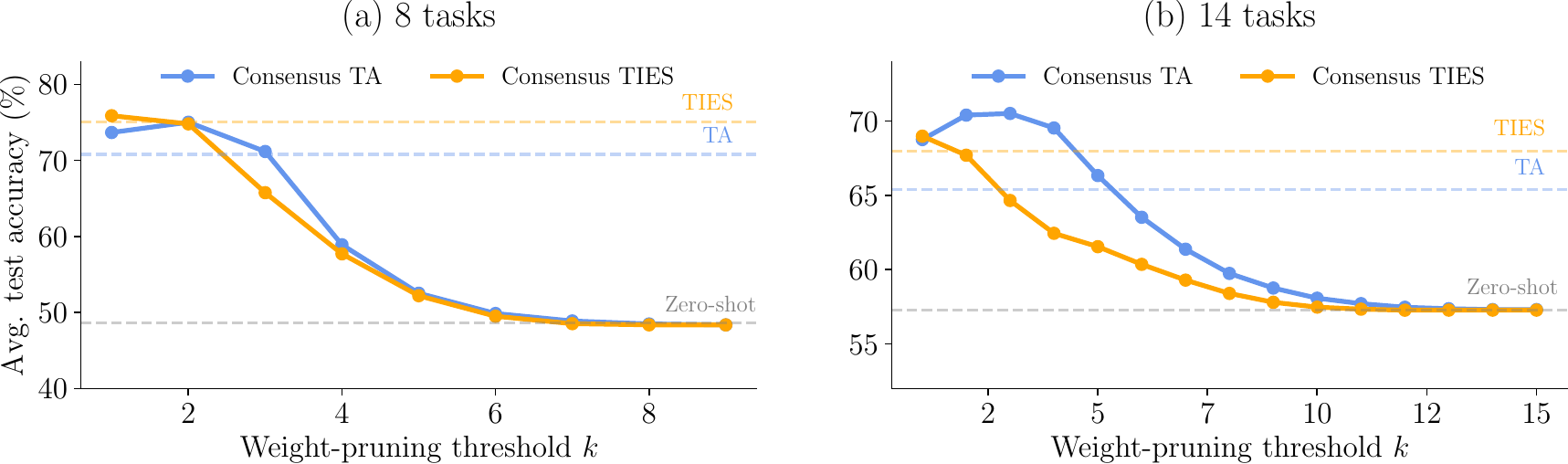}
    \caption{Performance of Consensus Merging with varying weight-pruning threshold $k$.}
    \label{fig:acc_vary_k_8_14task}
\end{figure*}

\subsection{Performance with Parameter-efficient fine-tuning methods}
\label{appendix:ia3 results}
In the main text we have presented the performance of our proposed methods for models with full fine-tuning. As Parameter-efficient fine-tuning (PEFT) methods have been widely adopted as cost-effective alternatives to full fine-tuning, in this section we provide the performance of our methods on PEFT models as well.

Following previous work \cite{Yadav_Tam_Choshen_etal_2023}, we provide the performance for models fine-tuned with (IA)$^3$ \citep{liu2022few} on the three NLP benchmarks. 

The results are presented in Table~\ref{tab:peft_nlp_results}. We observe from the results that while removing the selfish weights lead to performance degradation, possibly due to different profiles of the weights being tuned by (IA)$^3$ and full fine-tuning, removing only the catastrophic weights leads to consistent performance improvement of Consensus Merging over Task Arithmetic and TIES. For example, for the 7 NLP tasks, Consensus TA leads to 3.8\% gain to Task Arithmetic and Consensus TIES leads to 5.3\% gain to TIES.

\begin{table}[h]
\centering
\caption{Results on NLP benchmarks using (IA)$^3$ models, comparing the performance of Consensus Merging with baseline methods.}
\begin{tabular}{@{}cccccccc@{}}
\toprule
 &  &  & 7 NLP tasks &  & 8 QA tasks &  & All 11 tasks \\ \midrule
 &  &  & (IA)$^3$  &  & (IA)$^3$ &  & (IA)$^3$  \\ \midrule
\rowcolor[HTML]{EFEFEF} 
Zeroshot &  &  & 44.9 &  & 33.1 &  & 36.9 \\
Task arithmetic &  &  & \foo{67.1}{79.5} &  & \foo{57.6}{74.3} &  & \foo{59.7}{77.7} \\
Consensus TA $(k=1)$ &  &  & \foo{70.9}{83.6} &  & \textbf{\foo{60.2}{77.1}} &  & \textbf{\foo{60.8}{79.1}} \\
Consensus TA $(k=2)$ &  &  & \foo{66.6}{78.6} &  & \foo{54.1}{68.1} &  & \foo{55.7}{71.1} \\
TIES &  &  & \foo{66.0}{77.6} &  & \foo{56.8}{72.6} &  & \foo{56.1}{72.5} \\
Consensus TIES $(k=1)$ &  &  & \textbf{\foo{71.3}{84.2}} &  & \foo{58.7}{74.9} &  & \foo{59.2}{76.5} \\
Consensus TIES $(k=2)$ &  &  & \foo{67.6}{79.8} &  & \foo{56.3}{71.6} &  & \foo{53.1}{68.8} \\ \bottomrule
\end{tabular}
\label{tab:peft_nlp_results}
\end{table}

\subsection{Distribution of mask agreements with more tasks}
\label{app:selfish_weights_app}

We provide in Figure~\ref{fig:selfish weights app} the same histogram as in Figure~\ref{fig:selfish weights} for 14 tasks and 20 tasks, respectively for Task Arithmetic and TIES.

\begin{figure*}[t]
    \centering
    \includegraphics[width=1\linewidth]{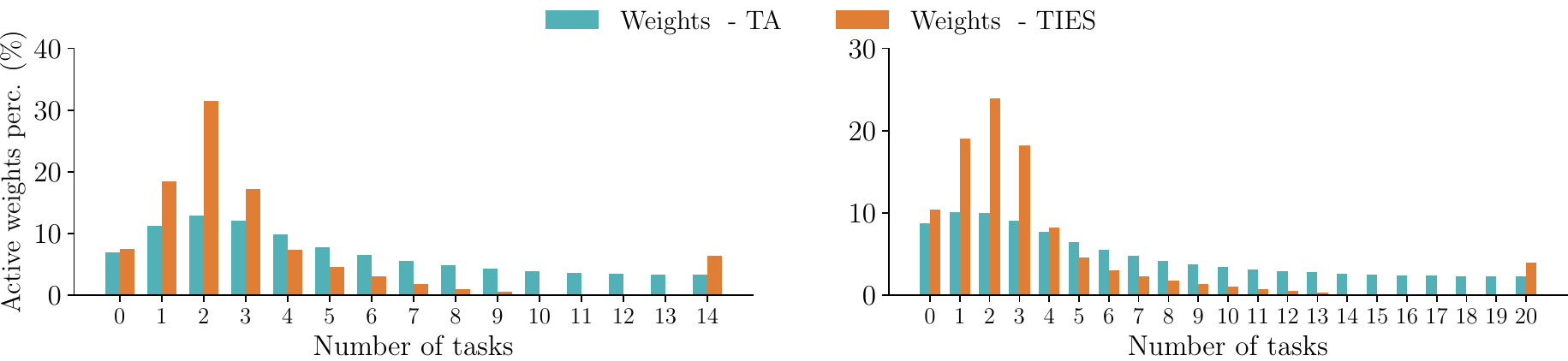}
    \caption{{The mask agreement profile, defined in \autoref{eq:mask agreement}, in the case of 14 and 20 vision tasks. This figure complements \autoref{fig:selfish weights}.}}
    \label{fig:selfish weights app}
\end{figure*}

\subsection{Result for ViT-B/16}
\label{app:result_vit_b_16}

We provide here the same results shown in \autoref{tab:vision_results} for ViT-B/16 in Table~\ref{tab:vision_results_vit_b_16}, where we observe similar findings as the case for ViT-B/32 and ViT-L/14 in main text.

\begin{table*}[ht]
\centering
\caption{Complementary to \autoref{tab:vision_results}, for results obtained with ViT-B/16.}
\renewcommand{\arraystretch}{1.5} 
\resizebox{0.8\textwidth}{!}{
\begin{tabular}{@{}ccccccccccc@{}}
\toprule
\multicolumn{2}{c}{} &  & \multicolumn{8}{c}{ViT-B/16} \\ \cmidrule(l){3-11} 
\multicolumn{2}{c}{} &  & \multicolumn{2}{c}{8 tasks} &  & \multicolumn{2}{c}{14 tasks} &  & \multicolumn{2}{c}{20 tasks} \\ \cmidrule(lr){4-5} \cmidrule(lr){7-8} \cmidrule(l){10-11} 
\multicolumn{2}{c}{\multirow{-3}{*}{Method}} &  & Acc.(\%)$\uparrow$ & Bits(Gb)$\downarrow$ &  & Acc.(\%)$\uparrow$ & Bits(Gb)$\downarrow$ &  & Acc.(\%)$\uparrow$ & Bits(Gb)$\downarrow$ \\ \midrule
\rowcolor[HTML]{EFEFEF} 
\cellcolor[HTML]{FFFFFF} & Zeroshot &  & 55.2 & {\color[HTML]{656565} 3.6} &  & 61.2 & {\color[HTML]{656565} 3.6} &  & 59.7 & {\color[HTML]{656565} 3.6} \\

\cellcolor[HTML]{FFFFFF} & Weight averaging &  & \foo{72.2}{76.6} & {\color[HTML]{656565} 3.6} &  & \foo{69.4}{74.7} & {\color[HTML]{656565} 3.6} &  & \foo{65.3}{70.3} & {\color[HTML]{656565} 3.6} \\

\cellcolor[HTML]{FFFFFF} & Task arithmetic &  & \foo{75.8}{80.2} & {\color[HTML]{656565} 3.6} &  & \foo{70.5}{75.8} & {\color[HTML]{656565} 3.6} &  & \foo{65.7}{70.7} & {\color[HTML]{656565} 3.6} \\
\cellcolor[HTML]{FFFFFF} & TIES &  & \textbf{\foo{79.7}{84.3}} & {\color[HTML]{656565} 3.6} &  & \foo{73.2}{78.7} & {\color[HTML]{656565} 3.6} &  & \foo{68.2}{73.3} & {\color[HTML]{656565} 3.6} \\

\cellcolor[HTML]{FFFFFF} & \textbf{Consensus Task arithmetic [ours]} &  & \foo{79.2}{83.6} & {\color[HTML]{656565} 3.6} &  & \textbf{\foo{74.3}{79.8}} & {\color[HTML]{656565} 3.6} &  & \textbf{\foo{69.7}{74.9}} & {\color[HTML]{656565} 3.6} \\

\multirow{-6}{*}{\cellcolor[HTML]{FFFFFF}\rotatebox[origin=c]{90}{Merging}} & \textbf{Consensus TIES [ours]} &  & \foo{79.4}{83.9} & {\color[HTML]{656565} 3.6} &  & \foo{74.1}{79.5} & {\color[HTML]{656565} 3.6} &  & \foo{68.7}{73.9} & {\color[HTML]{656565} 3.6} \\ \midrule
\rowcolor[HTML]{EFEFEF} 
\cellcolor[HTML]{FFFFFF}{\color[HTML]{333333} } & \cellcolor[HTML]{EFEFEF}{\color[HTML]{333333} Fine-tuned} & \cellcolor[HTML]{EFEFEF} & \cellcolor[HTML]{EFEFEF}94.6 & {\color[HTML]{656565} 22.9} &  & 92.8 & {\color[HTML]{656565} 39.4} &  & 93.2 & {\color[HTML]{656565} 56.0} \\
\cellcolor[HTML]{FFFFFF}{\color[HTML]{333333} } & Mag. Prunning &  & \foo{93.3}{98.5} & {\color[HTML]{656565} \textgreater7.0} &  & \foo{88.1}{94.7} & {\color[HTML]{656565} \textgreater7.5} &  & \foo{86.5}{92.7} & {\color[HTML]{656565} \textgreater8.1} \\
\cellcolor[HTML]{FFFFFF}{\color[HTML]{333333} } & Mag. Masking &  & \foo{89.7}{94.6} & {\color[HTML]{656565} 7.0} &  & \foo{84.8}{91.1} & {\color[HTML]{656565} 7.5} &  & \foo{81.6}{87.3} & {\color[HTML]{656565} 8.1} \\
\cellcolor[HTML]{FFFFFF}{\color[HTML]{333333} } & \textbf{TALL Mask + Task arithmetic} &  & \foo{94.2}{99.6} & {\color[HTML]{656565} 7.0} &  & \foo{92.0}{99.2} & {\color[HTML]{656565} 7.5} &  & \foo{92.5}{99.3} & {\color[HTML]{656565} 8.1} \\
\multirow{-5}{*}{\cellcolor[HTML]{FFFFFF}{\color[HTML]{333333} \rotatebox[origin=c]{90}{Compression}}} & \textbf{TALL Mask + TIES} &  & \textbf{\foo{94.6}{99.9}} & {\color[HTML]{656565} 7.0} &  & \textbf{\foo{92.6}{99.8}} & {\color[HTML]{656565} 7.5} &  & \textbf{\foo{93.0}{99.8}} & {\color[HTML]{656565} 8.1} \\ \bottomrule
\end{tabular}
}

\label{tab:vision_results_vit_b_16}
\end{table*}

\newcommand{\yes}{\checkmark}
\newcommand{\gt}{\textgreater}

\subsection{Full results on individual tasks}
\label{app:individual_accuracies}

\subsubsection{Full results for vision}

In main text we have provided the full results on individual tasks for ViT-B/32, here we provide the same radar plots for ViT-B/16 in Figure~\ref{fig:radar_plot_vitb16}, and the radar plots for ViT-L/14 in Figure~\ref{fig:radar_plot_vitl14}, respectively. We observe similar findings for the case with ViT-B/32 in main text.

\begin{figure*}[ht]
    \centering
    \includegraphics[width=1.0\linewidth]{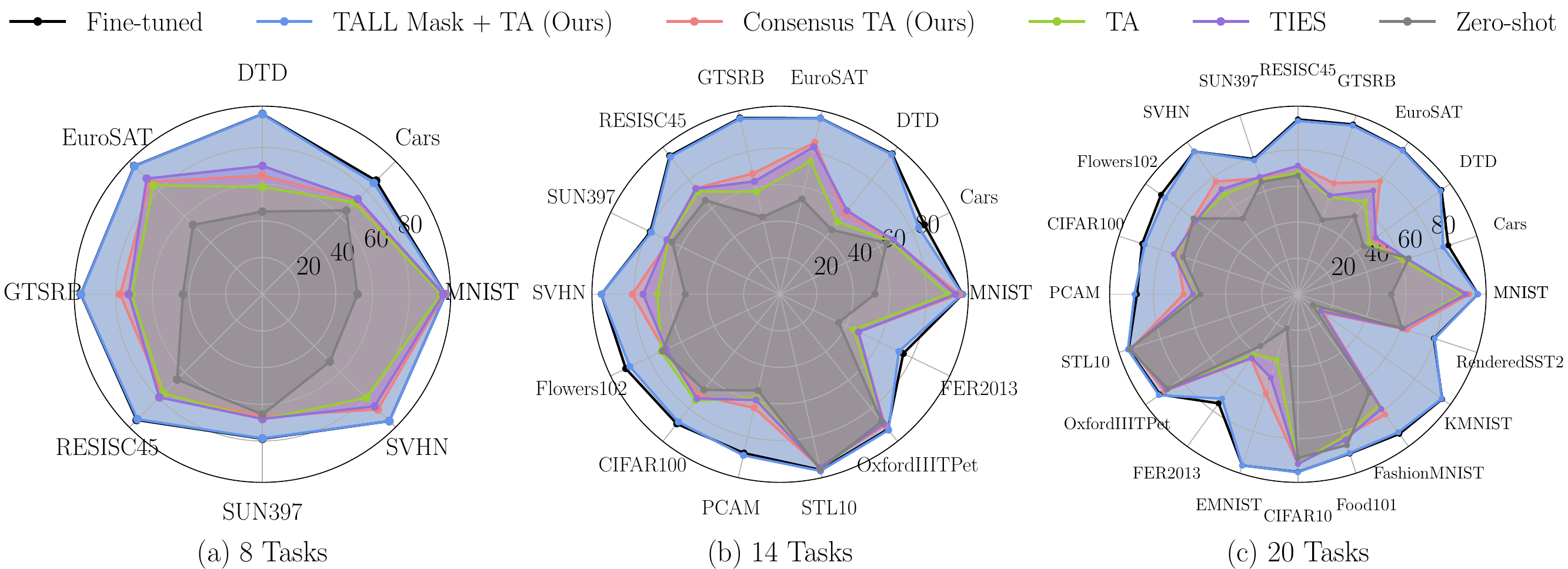}
    \caption{Absolute accuracy (\%) of individual tasks for ViT-B/16, comparing the accuracy of individual fine-tuned models, Our method, Task arithmetic, Ties-merging, and Zero-shot.}
    \label{fig:radar_plot_vitb16}
\end{figure*}

\begin{figure*}[ht]
    \centering
    \includegraphics[width=1.0\linewidth]{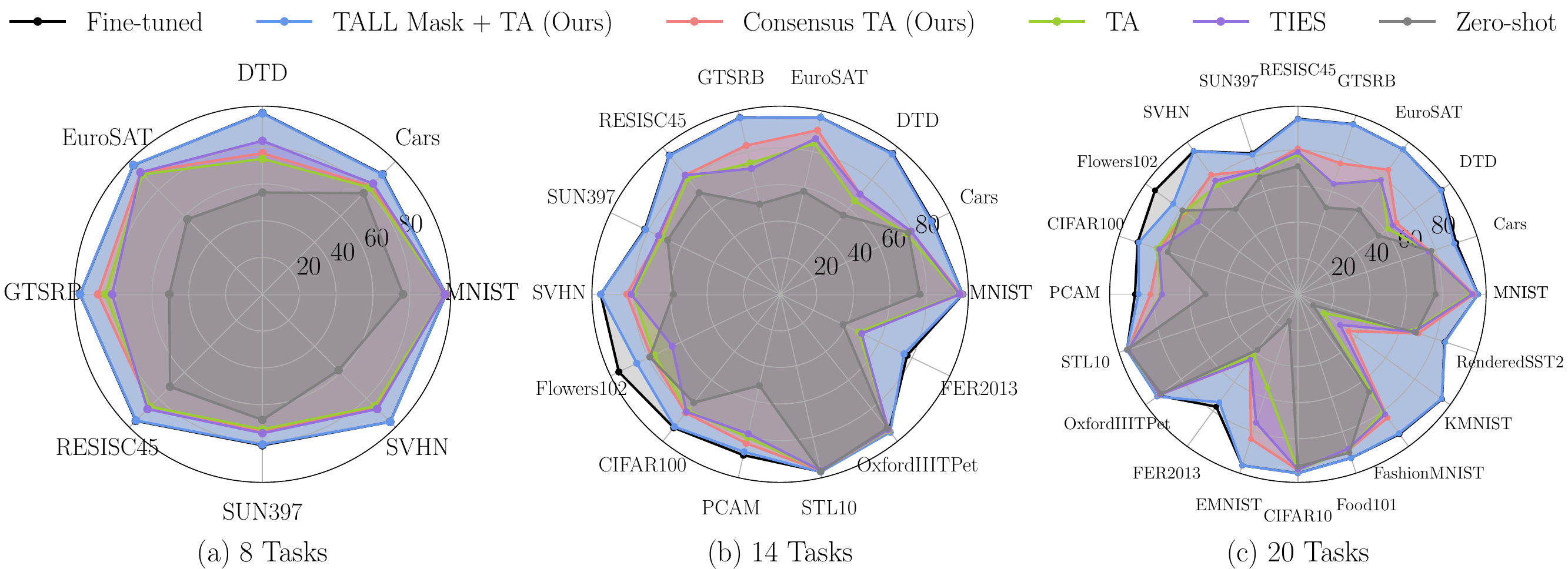}
    \caption{Absolute accuracy (\%) of individual tasks for ViT-L-14, comparing the accuracy of individual fine-tuned models, Our method, Task arithmetic, Ties-merging, and Zero-shot.}
    \label{fig:radar_plot_vitl14}
\end{figure*}

\subsubsection{Full results for NLP}

We provide the full results for individual tasks in NLP in Figure~\ref{fig:radar_plot_nlp}

\begin{figure*}[ht]
    \centering
    \includegraphics[width=1.0\linewidth]{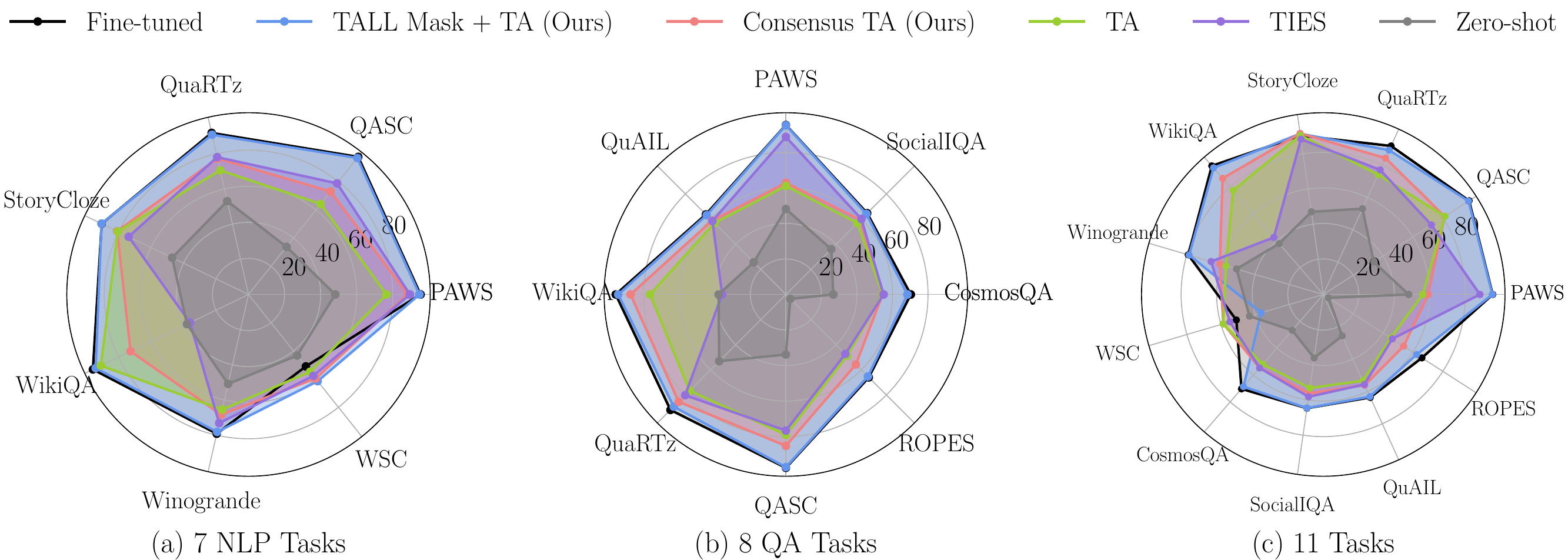}
    \caption{Absolute accuracy (\%) of individual tasks for T5-large on three benchmarks, comparing the accuracy of individual fine-tuned models, Our method, Task arithmetic, Ties-merging, and Zero-shot.}
    \label{fig:radar_plot_nlp}
\end{figure*}

\section{Sample subset selection protocol}
\label{app:sample_selection}

\autoref{fig:num_tasks vs acc and storage} presents the performance comparison as a function of different numbers of tasks. For a given number of tasks, we select 8 representative task combinations. Specifically, we sort all 20 tasks based on the following 8 orders:

\begin{itemize}
    \item Ascending order on zero-shot performance: 
       
       KMNIST, EMNIST, SVHN, GTSRB, FER2013, DTD, EuroSAT,
       MNIST, RenderedSST2, Cars, PCAM, RESISC45,
       FashionMNIST, SUN397, CIFAR100, Flowers102, Food101,
       OxfordIIITPet, CIFAR10, STL10
       
    \item Descending order on zero-shot performance: 
    
    STL10, CIFAR10, OxfordIIITPet, Food101, Flowers102, CIFAR100, SUN397,
       FashionMNIST, RESISC45, PCAM, Cars, RenderedSST2, MNIST,
       EuroSAT, DTD, FER2013, GTSRB, SVHN, EMNIST, KMNIST
    \item Wave order on zero-shot performance:

    Cars, PCAM, RenderedSST2, RESISC45, MNIST, FashionMNIST, EuroSAT, SUN397, DTD, CIFAR100, FER2013, Flowers102, GTSRB, Food101, SVHN, OxfordIIITPet, EMNIST, CIFAR10, KMNIST, STL10

    \item Zigzag order on zero-shot performance:
    
    STL10, KMNIST, CIFAR10, EMNIST, OxfordIIITPet, SVHN,
       Food101, GTSRB, Flowers102, FER2013, CIFAR100, DTD,
       SUN397, EuroSAT, FashionMNIST, MNIST, RESISC45,
       RenderedSST2, PCAM, Cars
    \item Ascending alphabetic order:

    CIFAR10, CIFAR100, Cars, DTD, EMNIST, EuroSAT, FER2013, FashionMNIST, Flowers102, Food101, GTSRB, KMNIST, MNIST, OxfordIIITPet, PCAM, RESISC45, RenderedSST2, STL10, SUN397, SVHN

    \item Descending alphabetic order:

    SVHN, SUN397, STL10, RenderedSST2, RESISC45, PCAM, OxfordIIITPet, MNIST, KMNIST, GTSRB, Food101, Flowers102, FashionMNIST, FER2013, EuroSAT, EMNIST, DTD, Cars, CIFAR100, CIFAR10

    \item Wave order on alphabetic order:

    CIFAR10, SVHN, CIFAR100, SUN397, Cars, STL10, DTD, RenderedSST2, EMNIST, RESISC45, EuroSAT, PCAM, FER2013, OxfordIIITPet, FashionMNIST, MNIST, Flowers102, KMNIST, Food101, GTSRB
    
    \item Zigzag order on alphabetic order:

    GTSRB, Food101, KMNIST, Flowers102, MNIST, FashionMNIST, OxfordIIITPet, FER2013, PCAM, EuroSAT, RESISC45, EMNIST, RenderedSST2, DTD, STL10, Cars, SUN397, CIFAR100, SVHN, CIFAR10
    
\end{itemize}

For a given number of tasks $n$, we retrieve the first $n$ tasks from these 8 sequences respectively, such that we account for both task difficulty (in zero-shot performance order) and randomness (in alphabetic order) when generating \autoref{fig:num_tasks vs acc and storage}.

\end{document}